\documentclass[10pt,twocolumn,letterpaper]{article}

\usepackage{iccv}
\usepackage{graphicx}
\usepackage{amsmath}
\usepackage{amssymb}
\usepackage{booktabs}
\usepackage{times}
\usepackage{epsfig}
\usepackage{subfig}
\usepackage{tabu}
\usepackage{multirow}
\usepackage[numbers]{natbib}
\usepackage{stfloats}
\usepackage{colortbl}
\usepackage{xcolor}
\definecolor{citecolor}{HTML}{0071bc}
\usepackage[pagebackref,breaklinks,colorlinks,
citecolor=citecolor,bookmarks=false]{hyperref}
\usepackage[capitalize]{cleveref}
\usepackage[accsupp]{axessibility}  

\newlength\savewidth\newcommand\shline{\noalign{\global\savewidth\arrayrulewidth\global\arrayrulewidth 1pt}\hline\noalign{\global\arrayrulewidth\savewidth}}

\iccvfinalcopy 

\def\iccvPaperID{8926}

\ificcvfinal\fi

\begin{document}

\title{Towards Fair and Comprehensive Comparisons for Image-Based \\ 3D Object Detection}

\author{Xinzhu Ma$^{1,2}$ \quad Yongtao Wang$^{3}$ \quad Yinmin Zhang$^{1,2}$ \quad Zhiyi Xia$^4$ \quad Yuan Meng$^{4,*}$ \\ Zhihui Wang$^{3,*}$ \qquad Haojie Li$^3$ \qquad Wanli Ouyang$^1$ \\ 
$^1$Shanghai AI Lab $^2$University of Sydney  $^3$Dalian University of Technology $^4$Tsinghua University \\
\texttt{\small\{maxinzhu, zhangyinmin, ouyangwanli\}@pjlab.org.cn}  \\ 
\texttt{\small\{yongtaowang@mail., zhwang@, hjli@\}dlut.edu.cn}  \quad \texttt{\small \{yuanmeng, xiazy21\}@mail.tsinghua.edu.cn}}

\maketitle
\pagestyle{empty}  
\thispagestyle{empty}

\begin{abstract}

In this work, we build a modular-designed codebase, formulate strong training recipes, design an error diagnosis toolbox, and discuss current methods for image-based 3D object detection. 
In particular, different from other highly mature tasks, e.g., 2D object detection, the community of image-based 3D object detection is still evolving, where methods often adopt different training recipes and tricks resulting in unfair evaluations and comparisons. 
What is worse, these tricks may overwhelm their proposed designs in performance, even leading to wrong conclusions. 
To address this issue, we build a module-designed codebase and formulate unified training standards for the community.
Furthermore, we also design an error diagnosis toolbox to measure the detailed characterization of detection models.  
Using these tools, we analyze current methods in-depth under varying settings and provide discussions for some open questions, e.g., discrepancies in conclusions on KITTI-3D and nuScenes datasets, which have led to different dominant methods for these datasets. 
We hope that this work will facilitate future research in image-based 3D object detection. 
Our codes will be released at \url{https://github.com/OpenGVLab/3dodi}.

\end{abstract}
\renewcommand{\thefootnote}{}
\footnote{Corresponding author}

\section{Introduction}
As a new and rapidly developing research field, vision-based 3D object detection \cite{3dodi_survey} shows promising potential in autonomous driving and attracts lots of attention from both academia and industry. Thanks to the unremitting efforts of numerous researchers, lots of advanced technologies, such as model designs \cite{m3drpn,gupnet,deviant}, detection pipelines \cite{centernet,pseudolidar,bevdet}, and challenging datasets \cite{kitti,nuscenes,waymo}, are continuously proposed, which significantly promotes the development of this research field.

However, although lots of breakthroughs in detection accuracy and inference speed have been achieved,  there are still some critical problems to be solved, especially the standardization of evaluation protocols. Specifically, compared with the encouraging developments at the technique level, the conventional rules in model building, training recipes, and evaluation are not well-defined. As shown in Table \ref{tab:overview}, existing methods generally adopt different settings, \eg backbones, training strategies, augmentations, \etc, to build and evaluate their models, and the commonly used benchmarks (and most of the papers) only record the final accuracy. 
This makes the comparison of the detectors unfair and may lead to misleading conclusions.

In this paper, we aim to provide a unified platform for image-based 3D object detectors and standardize the protocols in model building and evaluation. Specifically, similar to the advanced codebases in 2D detection \cite{mmdetection,detectron2}, we decompose the image-based 3D detection frameworks into several separate components, \eg backbones, necks, \etc, and provide a unified implementation for current methods.
Furthermore, we fully investigate existing algorithms and formulate several efficient training protocols, \eg training schedules, data augmentation, \etc. 
Our summarized training recipes not only provide a fair environment for evaluation but also significantly improve the performance of current methods, particularly in the KITTI-3D dataset.
For example,  the methods \cite{monodle,gupnet} published two years ago trained with our recipes can achieve better (or similar) performance than the recent works \cite{deviant,didm3d}, emphasizing the importance of establishing a standard training recipe. 

Furthermore, to systematically analyze the bottlenecks and issues in image-based 3D object detection, we urgently require a tool to thoroughly examine detection results.
Inspired by TIDE \cite{tide}, we propose an error diagnosis toolbox, named TIDE3D, to measure the detailed characterization of detection algorithms. Specifically, we decouple the detection errors into seven types, and then independently quantify the impact of each error type by calculating the overall performance improvement of the model after fixing the specified errors. In this way, we can analyze a specific aspect of given algorithms while isolating other factors.
In addition to characterizing the detailed features of the models, the proposed TIDE3D also has other useful applications. For example, the effectiveness and action mechanism of a specific design/module can be explored by analyzing the change of error distribution of the baseline model with or without the target design/module.

Besides, we also find the most concerned two datasets, \ie KITTI-3D \cite{kitti} 
 and nuScenes \cite{nuscenes}, are dominated by different detection pipelines, and even gradually become separate research communities. For example, none of the popular Bird's Eye View (BEV) detection methods provide the KITTI-3D results, and the top-performing methods in the KITTI-3D leaderboard also hard to achieve good results in nuScenes. We apply the cross-metric evaluation (\ie applying the nuScenes-style metrics on the KITTI-3D dataset and vice versa) on the representative models and report the adopted metric is the main factor causing this phenomenon. We also provide TIDE3D analyses for this issue.


To summarize, the contributions of this work are as follows:
First, we build a modular-designed codebase for the community of image-based 3D object detection, which can serve as a foundation for future research and algorithm implementation.
Second, we investigate the training settings and formulate standard training recipes for this task. 
Third, we provide an error diagnosis toolbox that can quantitatively analyze the detection models at a fine-grained level.
Last, we discuss some open problems in this field, which may provide insights for future research.
We hope our codebase, training recipes, error diagnosis toolbox, and discussions will promote better and more standardized research practices within the image-based 3D object detection community.

\begin{table*}[!t]
\centering
\resizebox{0.9\linewidth}{!}{
\setlength\tabcolsep{7.00pt}
\begin{tabular}{l|c|ccc|ccc}
\multirow{2}{*}{~} & \multirow{2}{*}{~} & \multicolumn{3}{c|}{KITTI} & \multicolumn{3}{c}{nuScenes} \\
~ & backbone & \# epochs  & data aug. & others & \# epochs  & data aug. & others \\ 
\shline
GUPNet \cite{gupnet} & DLA34-DLAUp & 140 & flip, crop & -  & -  & - & -   \\
FCOS3D \cite{fcos3d} & ResNet101-DCN-FPN & -  & -  & -  & 12  & flip & TTA   \\
PGD \cite{pgd} & ResNet101-DCN-FPN & 48  & flip  & TTA  & 24  & flip & TTA   \\
BEVDet \cite{bevdet} & ResNet50-FPN-LSS & -  & -  & -  & 24 & flip, BEV aug. & TTA  \\

\end{tabular}}
\caption{\textbf{Overview of example methods on KITTI-3D and nuScenes benchmarks.} The existing methods generally adopt different settings, \eg backbones, epochs, augmentations, \etc, in model training. Besides, KITTI-3D and nuScenes apply different evaluation metrics  and are dominated by different detection pipelines. TTA denotes the test-time augmentation.}
\label{tab:overview}
\end{table*}

\section{Related Work}

\noindent
{\bf Image-based 3D detection.} Image-based 3D detection is a rapidly developing research direction, and there are lots of detectors are proposed. Here we review these works based on the taxonomy proposed in \cite{3dodi_survey}: methods based on result-lifting, methods based on feature-lifting, and methods based on data-lifting. Refer to \cite{3dodi_survey} for the detailed and comprehensive literature review.

\noindent
{\it Result-lifting.} 
The methods in this branch first estimate the 2D projections and other items (\eg depth, orientation, \etc) of the 3D bounding boxes and then lift the results from the 2D image plane to the 3D world space.
In particular, pioneering works \cite{3dop,mono3d,m3drpn,centernet,fcos3d,stereorcnn,detr3d} introduce the popular 2D detection paradigms, \eg faster R-CNN \cite{fasterrcnn} and FCOS \cite{fcos}, into this research field, and lots of following works improve these baseline models in several aspects, including backbone designs \cite{m3drpn,deviant}, loss functions \cite{monodis,monodle,homoloss,roi10d}, depth estimation \cite{gupnet,monogeo,virtualdepth,pgd,densedepth,depthdeversity,iad3d}, geometric constraints \cite{monoground,geoconsistency,monojsg,monorun,smoke,monopsr,tlnet,gs3d,monogrnet}, feature embedding \cite{lineencoding,dimembedding}, key-point constrains \cite{deepmanta,rtm3d,autoshape}, depth augmented learning \cite{multi-fusion,d4lcn,monodistill,ligastereo,pseudostereo,graph_passing}, temporal sequences \cite{kin3d,dfm}, semi-supervised learning \cite{lpcg,pseudolabeling,semi-view,cmkd}, NMS \cite{shi2020distance,groomed}, data integration \cite{omni3d}, \etc.

\noindent
{\it Feature-lifting.} OFT \cite{oft} and DSGN \cite{dsgn} are the pioneering works in this research line, which lift the 2D features into 3D features (generally represented by BEV map) and then directly estimates the 3D bounding boxes with the resulting features. 
CaDDN \cite{caddn} uses a better transformation strategy \cite{liftsplat} and achieves promising results. BEVDet \cite{bevdet} introduces this paradigm into the multi-camera setting, and this BEV pipeline gradually dominates the nuScenes \cite{nuscenes} benchmark with effects of lots of follow-up works, such as \cite{bevformer,bevdepth,bevstereo}.

\noindent
{\it Data-lifting.} The data-lifting-based methods first estimate dense depth maps for the input images and then lift the pixels into 3D points using the estimated depth and camera parameters. After that, they generally leverage the LiDAR-based 3D detectors to predict the results from these `pseudo-LiDAR' signals \cite{pseudolidar,am3d}. The followers in this group are involved in the following aspects: improving the quality of pseudo-LiDAR \cite{pseudolidar++,wasserstein}, focusing on foreground objects \cite{taskaware,disprcnn,zoomnet}, end-to-ending training \cite{e2epl}, feature representation \cite{patchnet,is_needed}, geometric constraints \cite{monowithpl} or confidence refinement \cite{missconf}.

\renewcommand{\thefootnote}{$\dagger$}
\noindent
{\bf Codebases.} At present, there are two public codebases that support vision-based 3D object detection: MMDetection3D \cite{mmdet3d} and OpenPCDet \cite{openpcdet}. However, these two codebases are initially designed for LiDAR-based 3D object detection, and the supported vision-based methods are limited. In particular, MMDetection3D supports four vision-based detectors \cite{imvoxelnet,smoke,fcos3d,pgd}, and OpenPCDet only support CaDDN \cite{caddn}\footnote[1]{data collected at 24-02-2023}(also note some methods are developed based on these two codebases although they are not included in the official ones). In contrast, our codebase is designed for vision-based methods, excluding redundant codes and dependencies for LiDAR-based methods. 

\noindent
{\bf Error diagnosis for object detection.} To obtain the detailed characterization under the overall evaluation metric and show the strengths and weaknesses of the given models, some works~\cite{diagnosing, coco_analysis, tide} exist to analyze the errors in 2D detectors. In particular,  \cite{diagnosing} divides the errors of false positives into several categories and selects the top ${\rm N}$  most confident detections for errors analysis.  \cite{coco_analysis} further adds statistics for false negatives. More specifically, it progressively replaces the predicted items with corresponding ground-truth values and uses the $\Delta {\rm AP}$ metric to quantify the importance of each item.
\cite{tide} points out that this iterative approach incorrectly amplifies the impact of later error types, thus proposing a strategy where each error type is fixed independently. These works contribute to model analysis in the field of 2D detection, and to our best knowledge, there is no previous attempt to provide such a generic toolbox for 3D object detection. In this work, we provide a costumed error diagnosis toolbox to fill this gap.
\section{Codebase}
The objective of this work is to build a fair platform and provide experimental analysis for image-based 3D detection algorithms. For this purpose, we first build the general codebase which can support this work and facilitate future research.
Specifically, we decompose the image-based 3D detection frameworks into different components, \eg backbones, necks, heads, \etc, and we provide common choices for each component, \eg ResNet \cite{resnet}, DLA \cite{dla}, or DCN \cite{dcn2} for the backbones. 
In addition to the CNN modules, we also provide the implementation of other parts, such as augmentations, post-processing, \etc.
Currently, the proposed codebase supports more than ten methods such as \cite{monodle,gupnet,monodetr,monocon,smoke,centernet,fcos3d,pseudolabeling,lpcg,bevdet,pgd,monodistill} and two commonly used datasets \cite{kitti,nuscenes}. Our codebase will be publicly available and continuously maintained.

\noindent
{\bf Comparison with official implementation.} 
Here we provide a comparison of our implementation and the official implementation for some representative works.
In particular, here we take MonoDLE and GUPNet as examples for KITTI-3D dataset, and FCOS3D as example for nuScenes dataset.
Based on the results shwn in Table \ref{tab:implementation}, we can find our implementation gets consistent or better results than the official numbers. More details and the results for other models can be found in our public codebase.

\begin{table}[!t]
\centering
\resizebox{0.93\linewidth}{!}{
\begin{tabular}{l|ccc|cc}
\multirow{2}{*}{~}  & \multicolumn{3}{c|}{KITTI} & \multicolumn{2}{c}{nuScenes} \\
~ &  Easy & Mod. & Hard & mAP & NDS \\ 
\shline
MonoDLE \cite{monodle} & 17.89 &  13.87 & 12.03  & - & - \\
MonoDLE - ours  & 18.37 & 13.79 & 12.24 & - & - \\
GUPNet \cite{gupnet}& 21.48 & 15.22 & 12.79 & - & - \\
GUPNet - ours  & 21.66   &  15.45  & 12.51 & - & - \\
FCOS3D \cite{fcos3d}  & - & - & - & 30.6 & 38.1  \\
FCOS3D - ours  & -  & - & - & 30.4 & 38.5  \\
\end{tabular}}
\caption{{\bf Comparison of the implementations} on KITTI and nuScenes \emph{validation} sets. We report the average performance of the last epoch for both the official codes and our implementations over five runs. }
\label{tab:implementation}
\end{table}
\section{Diagnosis Toolkit}
\label{sec:tide3d}
We propose a general toolkit, named TIDE3D, to diagnose the cause of errors for 3D object detection models. We first briefly review the computing process of mean Average Precision (mAP) \cite{voc} and then introduce how to divide the detection errors into fine-grained types. Finally, we present how to weigh these errors and the implementation details.
We also recommend readers refer to TIDE \cite{tide} for preliminary knowledge.

\noindent
{\bf Review of mAP.}
As the most commonly used metric for evaluating object detection methods, mAP provides a comprehensive overview of a detector's performance.
Given the predictions and corresponding ground truths, each ground truth is matched to at most one prediction according to a specified metric, \eg KITTI-3D uses 3D Intersection over Union (IoU) and nuScenes adopts center distance as the metric. 
If multiple predictions meet the constraint, the ground truth only matches the prediction which has the maximum confidence score. 
The matched predictions are true positives (TP), and the remaining ones are false positives (FP).
After sorting the predictions by descending confidence, the number of true positives and false positives in the subset of predictions with confidence scores greater than $c$ is counted as ${\rm N}_{{\rm TP}}$ and ${\rm N}_{{\rm FP}}$. 
Then, the precision and recall of the predictions subset are obtained by:
\begin{equation} 
P_{c} = \frac{{\rm N}_{{\rm TP}}}{{\rm N}_{{\rm TP}}+{\rm N}_{{\rm FP}}} ~~~~~~~
R_{c} = \frac{{\rm N}_{{\rm TP}}}{{\rm N}_{{\rm GT}}}
\end{equation}
where ${\rm N}_{{\rm GT}}$ is the number of ground-truth objects. 
As the confidence threshold $c$ changes, the precision-recall curve is plotted and the Average Precision (AP) is obtained by calculating the area under the curve. Finally, mAP is obtained by averaging the AP values of all categories.

\noindent
{\bf Error definitions.}
Due to the complexity of calculating mAP, it prevents researchers from further analyzing the performance, and two detectors with the same mAP value may have different strengths and weaknesses. To address this issue, we segment the detection errors into several separate groups and measure the contribution of each error type. 
Specifically,  based on the taxonomy of 2D detection error proposed in TIDE \cite{tide} (including classification error, localization error, both classification and localization error, duplication error, background error, and missing error. See Table \ref{tab:errors} for details), we further propose two modifications for 3D object detection, including the sub-error types of localization  and the ranking error.

\noindent
{\bf Sub-error types of localization.} 
Since the localization of the 3D bounding box is determined by many factors, we further divide the localization error into three sub-errors to explore the effect of predicted location, dimension, and orientation on inaccurate localization.

\noindent
{\bf Ranking error.}  To describe each object in the scene, the detection model needs to output the 3D bounding box and its confidence. However, some detections may have higher confidence but lower accuracy. In the process of calculating AP, the detected boxes are sorted by \emph{descending confidence} and the precision of the high-confidence detections is calculated first. Therefore, the misalignment between confidence and the quality of the box naturally leads to the decline of AP and brings about possible room for improvement.

\noindent
{\bf Weighting the errors.} We use $\Delta {\rm AP}$ to quantify the impact of each error type. In particular, similar to TIDE \cite{tide}, we independently fix the errors (called ``oracle'', see Table~\ref{tab:errors} for the details), and compute the change of AP to measure the impact of this error type by:
\begin{equation} 
\Delta {\rm AP}_{o} = {\rm AP}_{o} - {\rm AP},
\end{equation}
where ${\rm AP}_{o}$ is the ${\rm AP}$ after applying the oracle $o$. In this way, we can capture the effect of each error type on the final metric, and a lower $\Delta {\rm AP}$ of the given error type indicates the model performs better at this part.

Due to current datasets generally adopting different evaluation metrics, the detailed implementations of TIDE3D are also slightly different for these datasets \cite{kitti,nuscenes,scannet}. More details of the specific implementations are provided in Appendix \ref{appendix:implementation}.

\begin{table*}[!t]
\centering
\resizebox{\linewidth}{!}{
\begin{tabular}{ m{3cm} | m{7cm} | m{8cm} }
\hline
Error Type & Definition & Oracle \\
\hline
Classification & 
Incorrect classification and correct localization.  & 
\multirow{2}{8.5cm}{Correct the category classification / location of detection, and delete this detection if it becomes duplicated. } \\
\cline{1-2}
Localization & Correct classification and incorrect localization. & \\
\cline{1-3}
Both Cls. and Loc. & Incorrect classification and incorrect localization.  & 
\multirow{3}{8cm}[-1.4em]{Delete the inaccurate / duplicate detection.}\\
\cline{1-2}
Duplication  & 
Correct classification and correct localization, but the corresponding GT has been matched by another higher-scoring detection. & \\
\cline{1-2}
Background & 
Detection from the background area.  & \\
\cline{1-3}
Missing & 
Undetected GT not covered by classification and localization error. &
Delete the missed ground truth.\\
\cline{1-3}
Ranking & 
Inconsistency between confidence and localization quality of detections. & Sort the detections by descending ${\rm IoU}$ score and calculate the precision of accurate detections first. \\
\hline
\end{tabular}}
\caption{{\bf Definition and oracle of detection errors.}}
\label{tab:errors}
\end{table*}
\section{Approach}

We first discuss the effects of common training settings on the final accuracy of current methods and then provide efficient and fair training recipes for further work. Furthermore, we also show the state-of-the-art (SOTA)
methods can be further promoted in performance with the summarized training recipes and some existing techniques in the KITTI-3D benchmark.
Besides, the detailed features of the detection algorithms are evaluated with the proposed TIDE3D, and we also show our diagnosis tool can be used to explore some open issues.
Finally, we discuss the conflicts between KITTI-3D and nuScenes datasets.

\noindent
{\bf Training recipes.} We observe that current methods generally adopt different training recipes, \eg the training schedule in KITTI-3D ranges from 60 epochs \cite{smoke} to 200 epochs \cite{didm3d}. We show that training differences like this greatly affect the performance of the algorithms and may mislead the latecomers to some extent, and it is necessary to formulate standard training recipes. Specifically, in KITTI-3D, we take the 70-epoch schedule as baseline (1$\times$) and also evaluate the models under 2$\times$ and 3$\times$ schedules. Similarly, the corresponding settings in nuScenes are 12, 24, and 36 epochs. We recommend the 2$\times$ schedule because this is the closest one to the common choice of other algorithms, and use other schedulers for further evaluation.
Furthermore, we also ablate other choices in model training and provide a general training recipe for image-based 3D detectors. We report, in addition to some basic choices such as random crop or horizontal flip, the photometric distortion and one-cycle learning rate \cite{onecycle} are also highly effective for our task and should be included in the conventional setting.

\noindent
{\bf Promoted baselines in KITTI-3D.} 
We also find, although KITTI-3D is released for a longer time and more methods are proposed on KITTI-3D benchmark, the baseline models in KITTI-3D are still \emph{under-tuned} compared to these on nuScenes.
Here we show current baseline models can be greatly promoted. We report that a two-year-ago model \cite{gupnet} (re-trained with pre-trained backbone, pseudo-labels, and the summarized training recipe) can achieve 17.65 $\rm{AP}_{40}$ on the KITTI-3D validation set, surpassing the SOTA by 1.11 points. Even so, we believe this number can be improved with further tuning or other tricks (\eg test-time augmentation and model-ensemble \cite{fcos3d}).

\noindent
{\bf Dataset and metrics.} KITTI-3D \cite{kitti} and nuScenes \cite{nuscenes} are the most concerned datasets in the image-based 3D detection field, however, they are dominated by different detection pipelines. We find that they are different in many aspects, and nuScene is generally considered to be more friendly to image-based algorithms. We adopt nuScenes-style evaluation on the KITTI-3D (and vice versa) and provide our discussion based on the cross-metric evaluation. More details are given in Section \ref{sec:kit_nus}.

\noindent
{\bf Detailed analysis with TIDE3D.}
With the TIDE3D introduced in Section \ref{sec:tide3d}, we can: (i) Analyze the detailed characteristics of 3D object detection models, whether it is an image-based model or LiDAR-based model,  evaluated with KITTI-3D-style or nuScene-style metrics, and designed for outdoor or indoor scenes (see Appendix \ref{appendix:more_tide3d_analysis} for the analyses for indoor dataset \cite{scannet} and LiDAR-based models).
 (ii) Conduct solid ablation studies and explore the action mechanisms for specific designs/modules.

\newcommand{\darr}[1]{$\texttt{#1}\downarrow$}
\newcommand{\uarr}[1]{$\texttt{#1}\uparrow$}

\begin{figure*}[!t]
\centering
\includegraphics[width=0.475\linewidth]{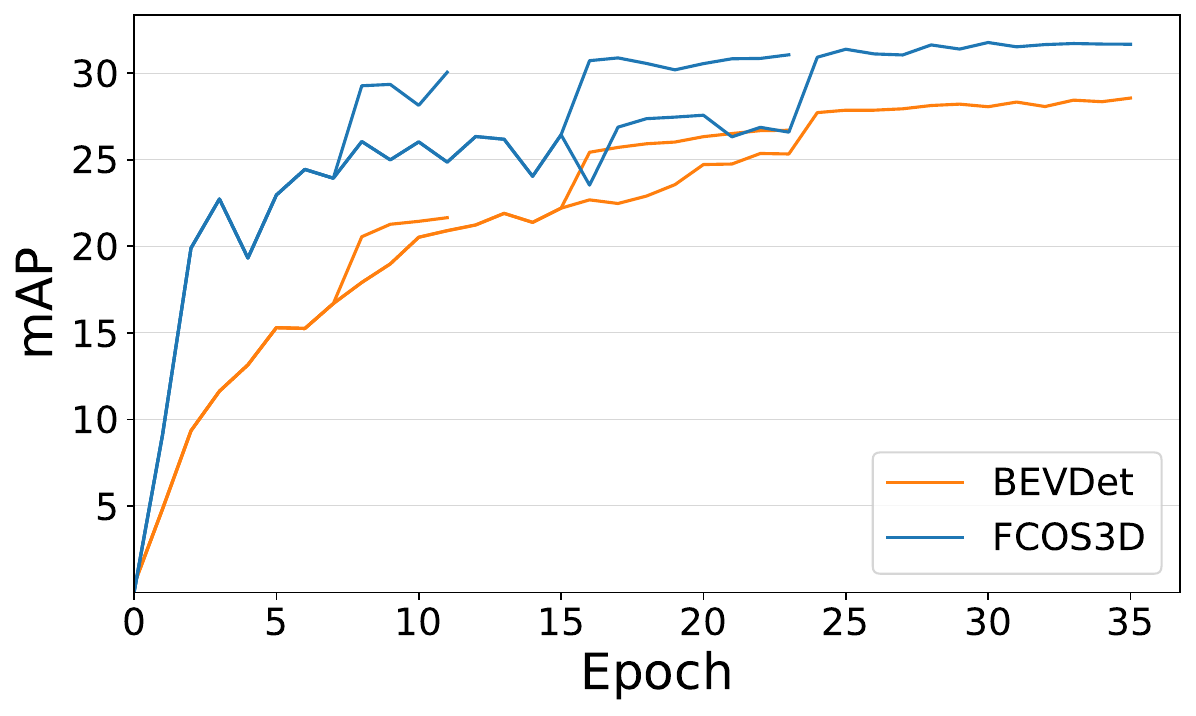}%
\includegraphics[width=0.475\linewidth]{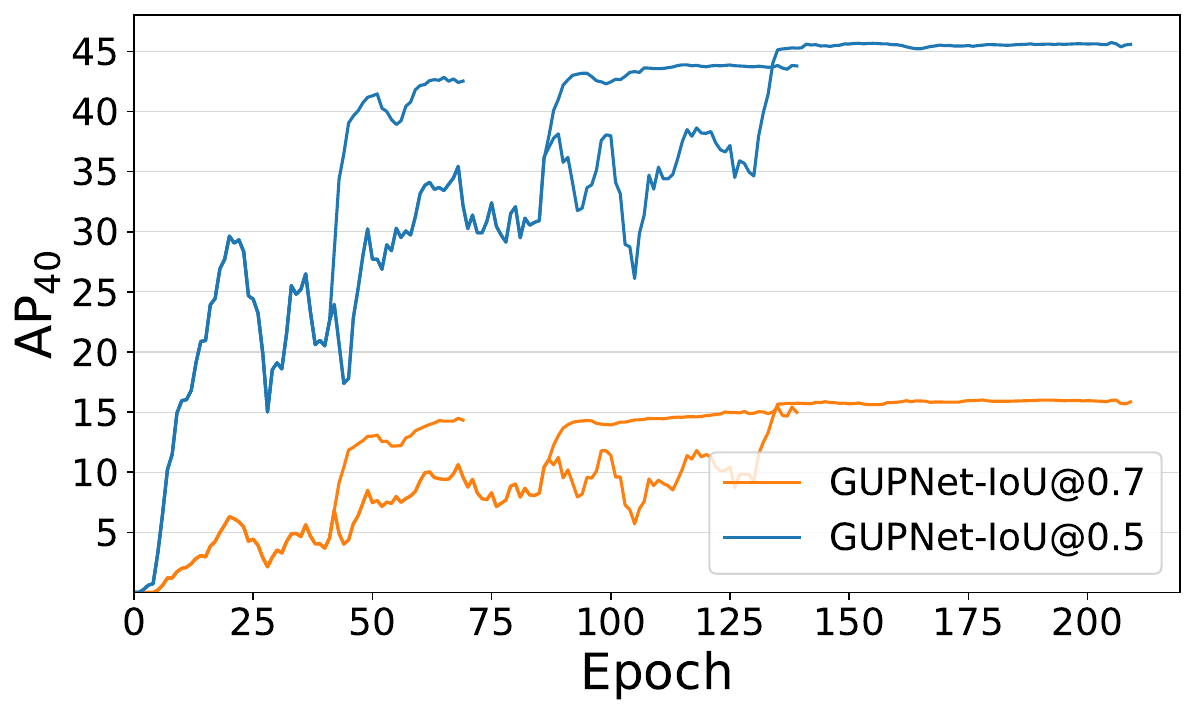}%
\vspace{-2ex}
\caption{
Learning curves on nuScenes ({\bf Left}) and KITTI-3D ({\bf Right}) with different detection models. With the 3$\times$ scheduler, BEVDet with R50-FPN optimized by AdamW and FCOS3D with R101-DCN-FPN optimized by SGD achieve 28.56 and 31.67 mAP on nuScenes, respectively, while GUPNet with DLA-34 optimized by Adam achieves 15.64 ${\rm AP}_{40}$@IoU=0.7 on KITTI-3D (\emph{validation} set). 
For better presentation, we present the curves by moving the average with windows size = 5. The original curves are shown in Appendix \ref{appendix:training_curve}.}
\label{fig:schedule_curve}
\end{figure*}

\section{Experiments and Analysis}
\noindent
{\bf Dataset and metrics.} We conduct experiments on KITTI-3D and nuScenes datasets, and 
 the detailed introduction for these two datasets is provided in Appendix \ref{appendix:dataset}.
Due to access restrictions of testing servers, the evaluation is conducted on \emph{validation} set, and we further apply cross-dataset metrics for better evaluation, \eg applying nuScenes-style metrics on the KITTI-3D dataset.

\noindent
{\bf Baseline models.} In this work, we mainly focus on the baselines based on the popular detection pipeline, including the CenterNet pipeline \cite{gupnet,deviant}, the BEV pipeline \cite{bevdet}, and the FCOS3D pipeline \cite{fcos3d,pgd}.

\noindent
{\bf Other settings.} To replicate the reported performance and minimize the impact of complex architectures and additional hyper-parameters, we adopt the simplest default configurations provided in the official code for most methods. Specifically, we use basic training schedules, simple architectures, and standard data augmentation techniques as the default settings. Further details beyond to these settings are provided in according sections.

\subsection{Training Recipes}

\noindent
{\bf Training schedules.} We first present the effects of training schedules. Specifically, we experiment with three different schedules, ranging from 1$\times$ to 3$\times$, for both KITTI-3D and nuScenes. 
Notably, we plot the training curves of all three schedules in the same figure to facilitate comparison. The AP curves shown in Figure \ref{fig:schedule_curve} indicate that schedules have a significant impact on the final performance, highlighting the importance of a standard scheduler for a fair comparison. Furthermore, the following observations can be summarized from these curves:

\noindent
{\bf i.} Different detection models require varying numbers of epochs as the optimal settings. 
For example, BEVDet exhibits significant and smooth performance improvements as the number of training epochs increases, while FCOS3D achieves near-optimal performance faster with perturbation after learning rate decay. This discrepancy may stem from the difference in optimizers, iterations, and detection pipelines. Intuitively, required training time can substantially vary across different settings, which may reduce the fairness of comparisons between them.

\noindent
{\bf ii.}
 Increasing the training procedure enhances the stability and precision of the fine-tuned model. Even after completing the 3$\times$ training, the trend of mAP/$\rm{AP}_{40}$ improvement continues to exist. The commonly accepted training schedule (mostly 1$\times$ or 2$\times$ on nuScenes) may be insufficient to unlock the full potential of a model. Therefore, the standard training procedure remains further discussed.

 \noindent
 {\bf iii.}
We observe a significant drop and a recovery at the beginning of the finetuning stage  of 1$\times$ training in the experiments of BEVDet and GUPNet. With training time increasing from 2$\times$ to 3$\times$, this phenomenon alleviates and even disappears. We suppose that longer training time helps the network learn more robust features, which prevents the overfitting brought by the sudden decay of the learning rate. More training steps at a small learning rate also help counter this issue.

\noindent
{\bf iv.} Even when the training is about to be completed, we still observe performances are unstable (see Appendix \ref{appendix:training_curve} for the un-smoothed curves). This phenomenon is particularly evident for the KITTI-3D-style metrics. Based on this, the comparison based on KITTI-3D metrics should be carefully conducted. 

\begin{table*}[!t]
\centering
\resizebox{0.8\linewidth}{!}{
\begin{tabular}{l|cc|ccc|ccc}
\multirow{2}{*}{~}  & \multirow{2}{*}{~} & \multirow{2}{*}{~} & \multicolumn{3}{c|}{${\rm AP}_{40}$@IoU=0.7} & \multicolumn{3}{c}{${\rm AP}_{40}$@IoU=0.5} \\
~ & schedule & pre-training & easy & mod. & hard & easy & mod. & hard  \\ 
\shline 
GUPNet \cite{gupnet}  & 2$\times$ & ImageNet & 21.48 & 15.22 & 12.79 & 59.30 & 43.83 & 38.40 \\ 
DID-M3D \cite{didm3d}  & $\approx$ 3$\times$ & ImageNet & 22.98 & 16.12 & 14.03 & - & - & - \\ 
DEVIANT \cite{deviant}  & 2$\times$ & ImageNet & 24.63 & 16.54 & 14.52 & 61.00 &  46.00 & 40.18 \\ 
\hline  
GUPNet  & 2$\times$  & None & 18.74 & 13.40 & 10.62 & 53.80 & 40.69 & 35.73 \\ 
GUPNet & 2$\times$  & ImageNet & 21.51 & 15.82 & 13.30 & 60.18 & 44.35 & 39.27 \\ 
GUPNet & 2$\times$  & DD3D & 24.25 & 15.82 & 13.26 & 63.16 & 44.60 & 40.08 \\ 
GUPNet + PL & 2$\times$  & DD3D & 24.71 & 17.25 & 15.21 & 63.73 & 46.56 & 42.45 \\ 
\end{tabular}}
\caption{
\textbf{Promoted baselines on KITTI-3D \emph{validation} set}. We report the 3D object detection performance of GUPNet with various pretrained backbones, evaluated by $\rm{AP}_{40}$@IoU=0.7/0.5. We also present SOTA methods for reference.
Note that both DID-M3D and DEVIANT are built based on the GUPNet baseline, and our modifications are only involved in the training phase, which shows the importance of building fair training recipes again.}
\label{tab:pretraining}
\end{table*}

\noindent
{\bf Other training recipes.} Due to strict geometric relations between the 2D image plane and the 3D world space, some complicated data augmentations are hard to be applied, and the horizontal flip is the common choice for existing methods. Note some simple geometric augmentations, like center crop and resize, are also commonly used in the KITTI-3D dataset, and see Appendix \ref{appendix:augmentation} for more discussions on such data augmentations. Here we show the photometric distortion and the one-cycle learning rate schedule are two cost-free training tricks to improve the detection accuracy (especially in KITTI-3D) and show their effectiveness with two representative baselines in Table \ref{tab:otherrecipes}.
We can find that both of them improve the GUPNet by a significant margin and can work together. Although nuScenes is a large-scale dataset, applying photometric distortion still brings a modest mAP improvement. Besides, we find the effects of one-cycle learning rate on nuScenes baselines are unstable, so we omit the corresponding cells to avoid misleading.

\begin{table}[h]
\centering
\resizebox{\linewidth}{!}{
\setlength\tabcolsep{4.00pt}
\begin{tabular}{l|cccc}
 ~ & baseline & w/ distortion & w/ one-cycle lr & full \\ 
\shline
GUPNet [$\rm{AP}_{40}$] & 15.26 & 16.54 & 15.87 & 16.83 \\
BEVDet [mAP] & 25.12 & 25.20 & - & - \\
\end{tabular}}
\caption{The photometric distortion and one-cycle learning rate are also effective training choices.}
\label{tab:otherrecipes}
\end{table}

\subsection{Promoted KITTI-3D Baselines.}

\noindent
{\bf Leveraging pre-training weights.}
We show the impact of weight initialization on detection models in Table \ref{tab:pretraining}. In particular, we compare three initial weights, including random initialization, ImageNet pretraining, and DD3D pertaining \cite{is_needed}. We can find that: {\bf i.} Similar to other computer vision tasks, model pre-training provides useful priors to the models and improve the performance significantly. {\bf ii.} As the data domains and proxy tasks become more similar (see Table \ref{tab:domain}), the pre-training weights also become more effective. Based on the above observations, designing custom pre-training algorithms, especially in an unsupervised manner, should be a promising research direction for future work.


\begin{table}[h]
\centering
\resizebox{\linewidth}{!}{
\setlength\tabcolsep{7.00pt}
\begin{tabular}{r|ccc}
 ~ & domain & proxy task & \# img. $\times$ \# ep. \\ 
\shline
ImageNet & daily scenes & classification & 1.3M $\times$ 100 \\
DD3D & driving scenes & depth estimation& 15M $\times$ 12.8 \\
\end{tabular}}
\caption{Information of different pre-training models. Ep. denotes the pretraining epochs.}
\label{tab:domain}
\end{table}

\noindent
{\bf Pseudo-labeling.} According to \cite{lpcg,pseudolabeling}, we generate the pseudo-labels (only for the 3,712 training images) from a LiDAR-based model \cite{pvrcnn}. Then we use them to train GUPNet and report the performance in Table \ref{tab:pretraining}. According to the results, the final result of our GUPNet can achieve 17.65 $\rm{AP}_{40}$ under the moderate setting, surpassing the previous SOTA (DEVIANT) by 0.71 points. We emphasize that we do not make any changes to the detection model, and the modifications are only involved in the training phase. This result shows the urgency and importance of establishing a fair and efficient training recipe.

\subsection{TIDE3D Analysis}
\label{sec:tide_analysis}

\begin{figure*}[!t]
\centering
\subfloat[KITTI dataset]{\includegraphics[width=3.3in]{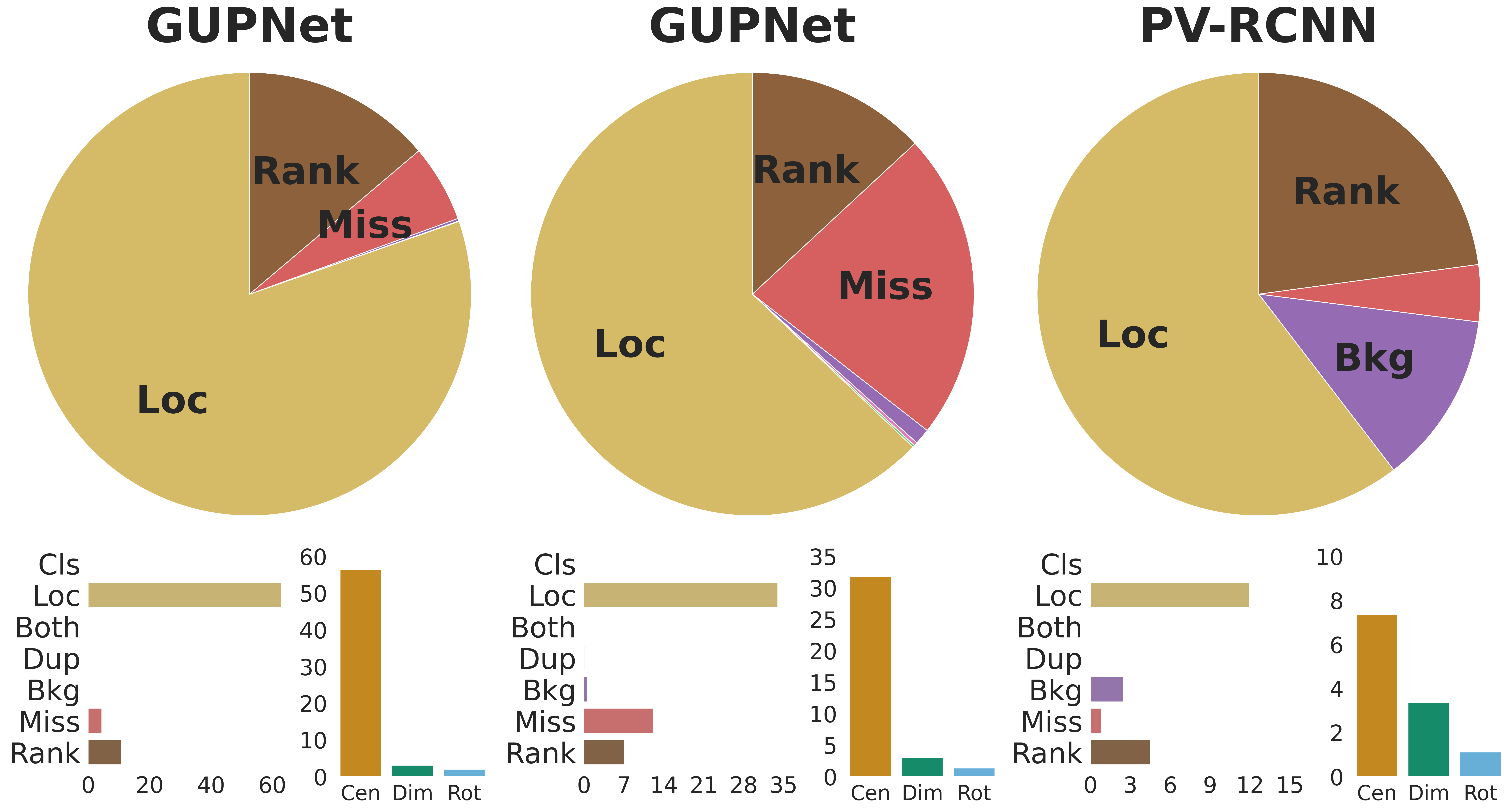}%
\label{fig:kitti_car}}
\hfill{\color{black}\vrule}\hfill
\subfloat[nuScenes dataset]{\includegraphics[width=3.3in]{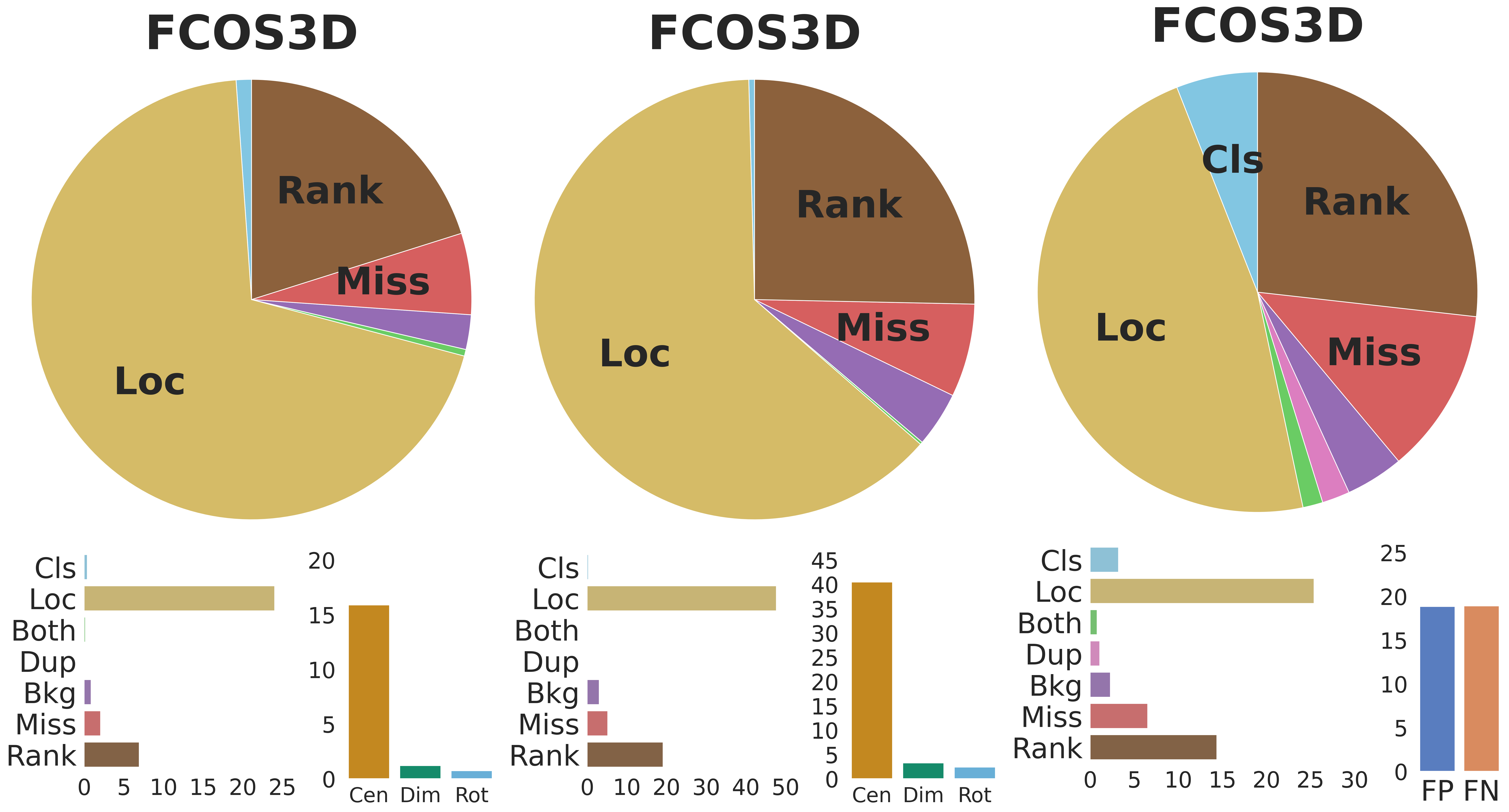}%
\label{fig:nuscenes}}
\caption{{\bf Example error diagnosis results.} 
(a) Error analyses on KITTI-3D \emph{validation} set under the moderate setting.
From left to right, we show the results of GUPNet \cite{gupnet} for the Car category with 0.7 IoU threshold, 0.5 IoU threshold, and the results of a LiDAR-based method \cite{pvrcnn} for the Car category with 0.7 IoU threshold. (b) Error diagnosis results of FCOS3D \cite{fcos3d} on nuScenes \emph{validation} set. From left to right, we show error diagnosis results based on KITTI-3D-style metrics with IoU@0.5 for all categories, the Car category, and the results based on nuScenes-style metrics for all categories.}
\label{fig:tide3d_summary}
\end{figure*}

\noindent
{\bf Error diagnosis.}
Here we present the error diagnosis results of some representative methods, settings, and evaluation metrics in Figure \ref{fig:tide3d_summary}. According to these error distributions, we can get the following observations:

\noindent
{\bf i.} Similar to \cite{monodle,deviant,is_needed,pgd}, we find localization error is the bottleneck of current methods and it is mainly caused by the inaccurate location (center). Even for the LiDAR-based method \cite{pvrcnn} which estimates the results from the data with accurate spatial information, localization error still ranks first among all error types. Differently, for \cite{pvrcnn}, the dimension and the orientation also have a significant impact on the quality of localization. 

\noindent
{\bf ii.} We find the misalignment between the confidence and the bounding box's quality is a serious problem in this field, suggesting that giving more accurate confidence is an effective way to improve detection accuracy. An empirical study of this problem is given in the following part.

\noindent
{\bf iii.} Under a lower IoU threshold, the KITTI-3D-style metric is still dominated by the localization error. Meanwhile, this issue has been alleviated on nuScenes-style metric. This indicates 3D IoU-based AP is a very sensitive metric, and center distance-based AP is more friendly to image-based methods. More TIDE3D results and analyses, \eg indoor scenes, are provided in Appendix \ref{appendix:more_tide3d_analysis}.

\noindent
{\bf Validating design choices.} 
We show the proposed TIDE3D can validate whether a given design choice supports its motivation/claim. For instance, some works, such as \cite{gupnet,missconf,omni3d} proposed their designs to get better confidence scores of detection results.
Here we use TIDE3D to validate the design of GUPNet.
Particularly, GUPNet models the uncertainty of estimated depth ($\sigma$) and further gets the confidence of depth with this uncertainty by: $p_{depth} = exp(-\sigma)$. Finally, the 3D confidence ($p_{3D}$) can be obtained by correcting 2D confidence ($p_{2D}$) with depth confidence: $p_{3D} = p_{depth} \cdot p_{2D}$. Table \ref{tab:tide3d_gupnet_score} gives the results of GUPNet equipped with 2D confidence and 3D confidence respectively. We observe that the $\rm{AP}_{40}$ improved by 2.21 and the ranking error reduces to 10.81 from 15.10, which confirms the effectiveness of this design. Meanwhile, we find the localization error also reduced, which is caused by the high-quality results being improved and corresponding results having a higher priority in AP computing. See Appendix \ref{appendix:more} for more analysis about the ranking error and localization error in TIDE3D.

\begin{table}[!t]
\centering

\newcommand{\dmAP}[1]{\text{E}_\texttt{#1}\downarrow}

\resizebox{\linewidth}{!}{
\begin{tabular}{l|ccccc}
\multirow{2}{*}{~} 
~ &  ${\rm AP}_{40}\uparrow$  & $\dmAP{loc}$ & $\dmAP{bkg}$ & $\dmAP{miss}$ & $\dmAP{rank}$  \\ 
\shline
GUPNet w/ 2D conf.  & 14.81 & 69.87 & 0.46 & 1.73 & 15.10  \\ 
GUPNet w/ 3D conf.  & 17.02 & 62.84 & 0.15 & 4.41 & 10.81  \\ 
Improvement & +2.21 & -7.03 & -0.31 & +2.68 & -4.29 \\ 
\end{tabular}}
\caption{{\bf Effects of 3D confidence.} Experiments are conducted on KITTI-3D \emph{validation} set. $\Delta {\rm AP}_{40}$ is denoted as E for brevity.}
\label{tab:tide3d_gupnet_score}
\end{table}





\begin{table*}[!t]
\centering
\setlength\tabcolsep{8.0pt}
\resizebox{0.71\linewidth}{!}{
\begin{tabular}{l|c|c|ccc|cc}
\multirow{2}{*}{~} &\multirow{2}{*}{Dataset} &\multirow{2}{*}{Class} & \multicolumn{3}{c|}{KITTI Metric} & \multicolumn{2}{c}{nuScenes Metric} \\
\cline{4-8}
~ & & & @0.7 & @0.5 & @0.25 & ${\rm mAP}$ & ${\rm NDS}$  \\ 
\shline
GUPNet & KITTI-3D & Car & 14.85 & 40.85 & 62.87 & 47.57 & 61.85 \\ 
PV-RCNN & KITTI-3D & Car & 82.41 & 94.31 & 94.47 & 92.74 & 92.61 \\ 
FCOS3D & nuScenes & Car & 1.83 & 20.87 & 50.74 & 47.96 & 54.32 \\
FCOS3D & nuScenes & All  & 0.26 & 6.06 & 17.57 & 31.05 & 40.31 \\ 
BEVDet & nuScenes & All & 0.10 & 6.55 & 19.25 & 30.75 & 38.22 \\ 
\end{tabular}}
\caption{
\textbf{Evaluation Metric.} 
We exchange the metrics between KITTI-3D and nuScenes for the evaluation of 3D object detectors.
This provides a standardized way to compare the performance of 3D object detectors across the two datasets. 
Dataset represents the source of training data for the detector. For detectors trained on KITTI-3D, we only compute translation, scale, and orientation for NDS under nuScenes metric. For KITTI metric, we evaluate the results of cars under the hard strict with varying IoU thresholds.
}
\label{tab:metric}
\end{table*}

\begin{figure}[!t]
\centering
\subfloat{\includegraphics[width=1.45in]{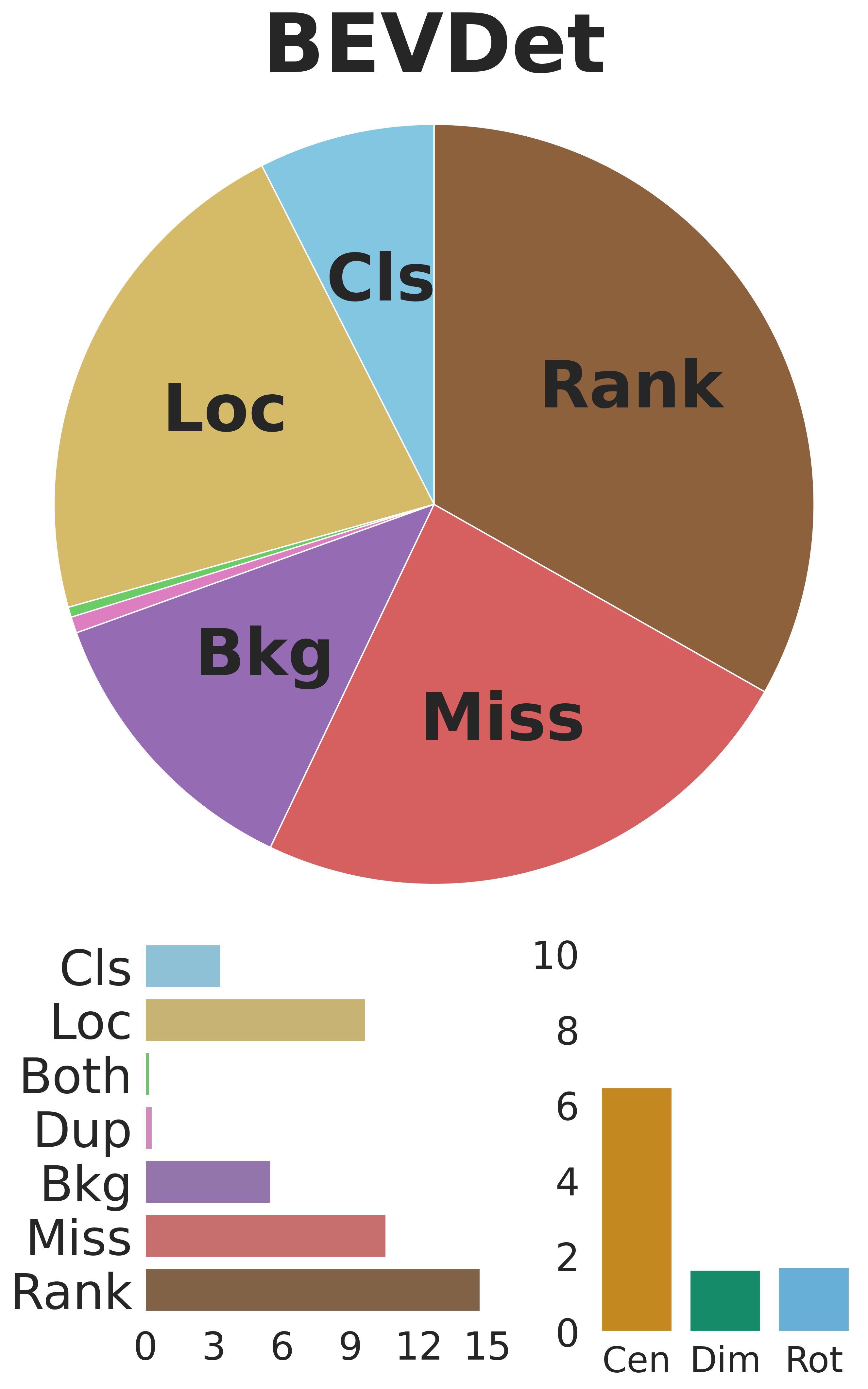}%
\label{fig:bevdet_iou0.25}}
\hfill
\subfloat{\includegraphics[width=1.45in]{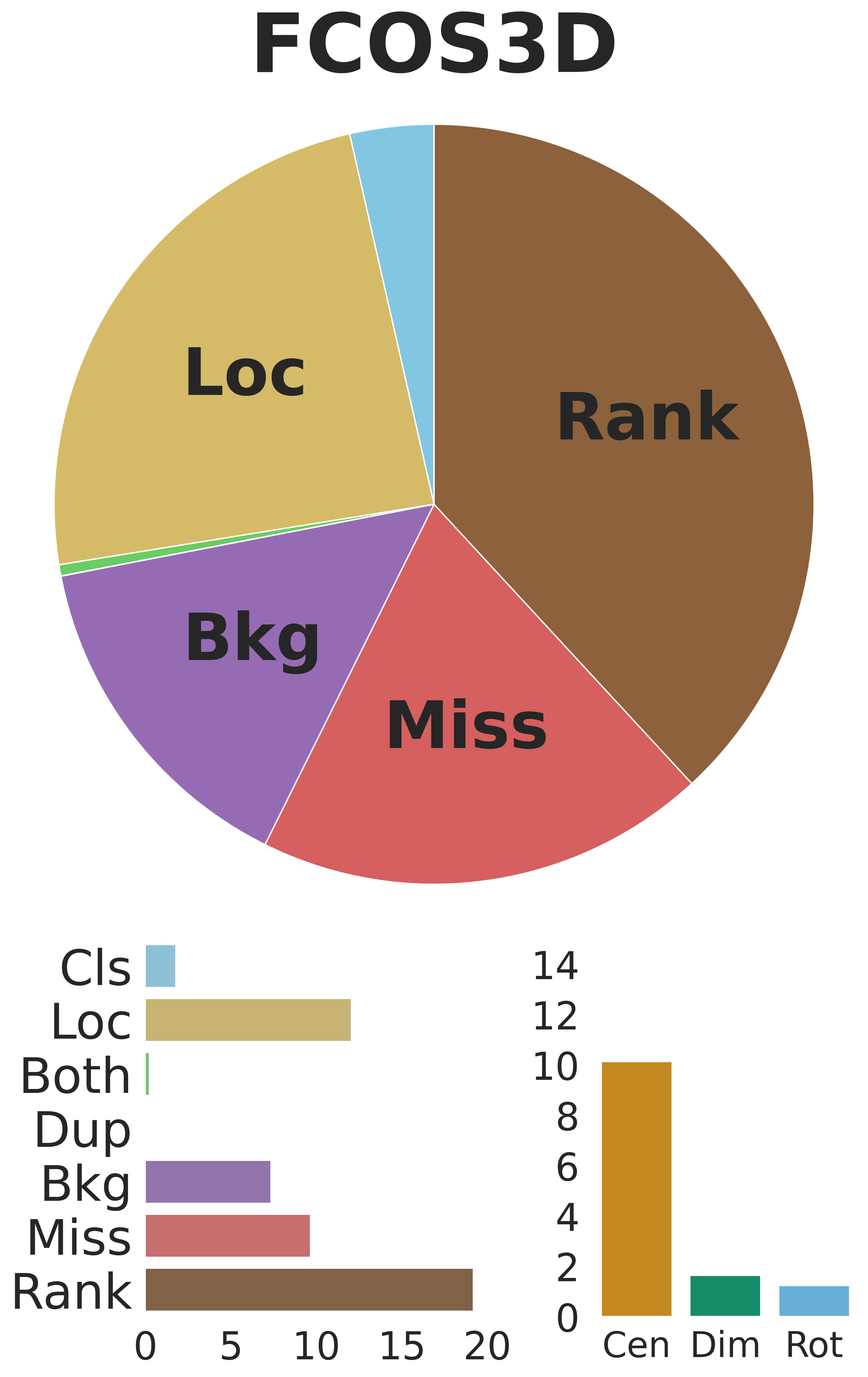}%
\label{fig:fcos_iou0.25}}
\caption{
\label{fig:nuscenesv2}
{\bf TIDE3D analyses} of BEVDet \cite{bevdet} and FCOS3D \cite{fcos3d} for all categories on nuScenes \emph{validation} set based on the KITTI-style metric with 0.25 IoU threshold. Our results indicate that FCOS3D is impacted by various errors, while BEVDet is mainly dominated by localization errors.}
\vspace{-5pt}
\end{figure}

\subsection{Metric Analysis for KITTI-3D and nuScenes}
\label{sec:kit_nus}

Additionally, we have observed interesting discrepancies in evaluation metrics between KITTI-3D and nuScenes. 
Specifically, as shown in Table \ref{tab:metric}, GUPNet and FCOS3D get similar AP of Car category under the nuScenes-style metric, but they show a significant performance gap under the KITTI-style metric. As we know, KITTI-3D primarily uses the strict 3D IoU metric with a certain threshold as the metric, while nuScenes measures the center distance and a series of decoupled error attributes such as translation, scale, orientation, \etc, separately (Appendix \ref{appendix:dataset} provides the details of metrics).
We find that translation, scale, and orientation jointly affect the result of IoU, which means that methods achieving competitive performance on the KITTI-3D metric require comprehensive development, while methods with special strengths in nuScenes can also get a good performance. According to our experience, a model with good performance under the KITTI-style metric generally performs well under the nuScenes-style metric, but the opposite is not always true.

Furthermore, we present TIDE3D analyses of BEVDet \cite{bevdet} and FCOS3D \cite{fcos3d} for all categories on nuScenes \emph{validation} set based on the KITTI-style metric with 0.25 IoU threshold in Figure~\ref{fig:nuscenesv2}. Based on the distribution of localization sub-errors (6.5-1.6-1.7 {\it v.s.} 10.1-1.7-1.2), we can find that FCOS3D makes more accurate predictions for orientation and dimension than BEVDet.
One of the reasons for this is the loss of height information of BEVDet in the LSS (lift, splat, shoot) process, which serves to transform the image feature into BEV space. Consequently, the model may struggle to estimate the height of objects accurately and the center height of objects.
See Appendix \ref{appendix:ahe} for more results and analyses.


The above two reasons explain why the mainstream methods for KITTI and nuScenes are distinct, and it is reasonable and necessary to separately design models to match the preferences of these two kinds of metrics.
\vspace{6pt}
\section{Conclusion}
In this paper, we aim to standardize the evaluation protocols for image-based 3D object detection, an ever-evolving research field. We build a modular-designed codebase and provide efficient training recipes for this task, leading to significant improvements in the performance of current methods. Besides, we offer an error diagnosis toolbox to measure the detailed characterization of detection algorithms and highlight some open problems in the field. We hope our codebase, training recipes, error diagnosis toolbox, and discussions will foster better and more standardized research practices within the image-based 3D object detection community. 

\section*{Acknowledgments}
This work was supported by the Australian Medical Research Future Fund MRFAI000085, CRC-P Smart Material Recovery Facility (SMRF) – Curby Soft Plastics, CRC-P ARIA - Bionic Visual-Spatial Prosthesis for the Blind, and National Natural Science Foundation of
China (NSFC Grants NO. 61976038).

{\small
\bibliographystyle{ieee_fullname}
\bibliography{references}
}

\clearpage
\newpage
\setcounter{section}{0}
\renewcommand\thesection{\Alph{section}} 

\section{Appendix}

\subsection{Datasets and Metrics}
\label{appendix:dataset}

\noindent
{\bf KITTI-3D.}
KITTI \cite{kitti} is the most popular dataset in autonomous driving scenario in the past decade. KITTI-3D is a subset of KITTI dataset, which contains 7,481 and 7,518 frames for training and testing respectively. In each frame, the stereo images, synchronized LiDAR sweep, calibration files, and 3D bounding box annotations are provided. Following \cite{3dop,mv3d}, we split the training frames into a training set (3,712 frames) and a validation set (3,769 frames), and conduct the experiments in this split. 
Following \cite{monodis}, we adopt the ${\rm AP}_{40}$ of 3D detection and Bird's Eye View (BEV) detection as metrics. We mainly focus on the Car category on KITTI-3D in the main paper, and both 0.7 and 0.5
IoU thresholds are considered. In this supplementary, we also discuss the Pedestrian and Cyclist categories with 0.5 IoU threshold.

\noindent
{\bf nuScenes.} nuScenes \cite{nuscenes} is a large-scale autonomous driving dataset proposed in 2020.
It provides about 40K annotated frames of 10 categories for the panoramic view in the autonomous driving scenario. In particular, there are 28,130 frames, 6,019 frames, and 6,008 frames for training, validation, and testing respectively (six images per frame). As for evaluation metrics, nuScenes uses a  2D center error (in the ground plane) based AP metric. Specifically, they consider four thresholds, $\mathbb{D}=\{0.5,1,2,4\}$, for AP computing, and average them over 10 categories to get the final mAP:
\begin{equation}
{\rm mAP} = \frac{1}{|\mathbb{C}||\mathbb{D}|}\sum_{c \in \mathbb{C}}\sum_{d\in \mathbb{D}}{\rm AP}_{c, d},
\label{equ:map_nuscenes}
\end{equation}
where $|\mathbb{C}|$ and $|\mathbb{D}|$ are the category set and threshold set respectively.
Furthermore, five true positive metrics (TP metrics) are also considered, including Average Translation Error (ATE), Average Scale Error (ASE), Average Orientation Error (AOE), Average Velocity Error (AVE), and Average Attribute Error (AAE). For each TP metric, the mean TP metric (mTP metric) are computed by:
\begin{equation}
{\rm mTP}_k = \frac{1}{|\mathbb{C}|} \sum_{c \in \mathbb{C}} \mathrm{TP}_{k, c},
\label{equ:mean_tp_metrics}
\end{equation}
where $k$ is the index of TP metrics, \eg  $\mathrm{TP}_{1}$ is ATE.
Finally, the nuScenes detection score (NDS) is computed by weighted averaging the above metrics:
\begin{equation}
{\rm NDS} = \frac{1}{10} [5\cdot {\rm mAP}+\sum_{k=1}^5(1-\min(1, {\rm mTP}_{k}))].
\label{equ:mean_tp_metrics2}
\end{equation}

\noindent
{\bf ScanNet V2.} ScanNet V2 \cite{scannet} is a richly annotated dataset of 3D reconstructed of indoor scenes, and 3D box annotations can be derived from the annotated meshes \cite{votenet}. There are about 1.2K training samples with 18 object different categories for hundreds of rooms. In this dataset, we adopt the 3D IoU based mAP as metric, and 0.25 and 0.5 IoU thresholds are considered in model evaluation and error analysis.

\subsection{Average Height Error}
\label{appendix:ahe}
In the main paper, Table \ref{tab:metric} shows that different evaluation metrics have their preferences for models. Specifically, on the nuScenes dataset, FCOS3D and BEVDet show similar performances under the nuScenes metric, but there is a large gap under the KITTI metric. 
To further explore this phenomenon, we define Average Height Error (AHE) as the 1D Euclidean distance of the center height of the bounding box, in order to bridge the gap between 2D IoU and 3D IoU.
Table \ref{tab:center_height} shows that the predicted center height of BEVDet is significantly worse than that of FCOS3D. 
It reflects that the nuScenes metric is more friendly to BEV-based methods.

\begin{table}[h]
\centering
\resizebox{0.5\linewidth}{!}{
\setlength\tabcolsep{7.00pt}
\begin{tabular}{l|cc}
 ~ & mATE & mAHE \\ 
\shline
FCOS3D & 0.78 & 0.11 \\
BEVDet & 0.72 & 0.15 \\
\end{tabular}
}
\caption{Average Translation Error (ATE) and Average Height Error (AHE) of BEVDet and FCOS3D for all categories on nuScenes \emph{validation} set.}
\label{tab:center_height}
\end{table}

\begin{figure}[!t]
\centering
\includegraphics[width=1.0\linewidth]{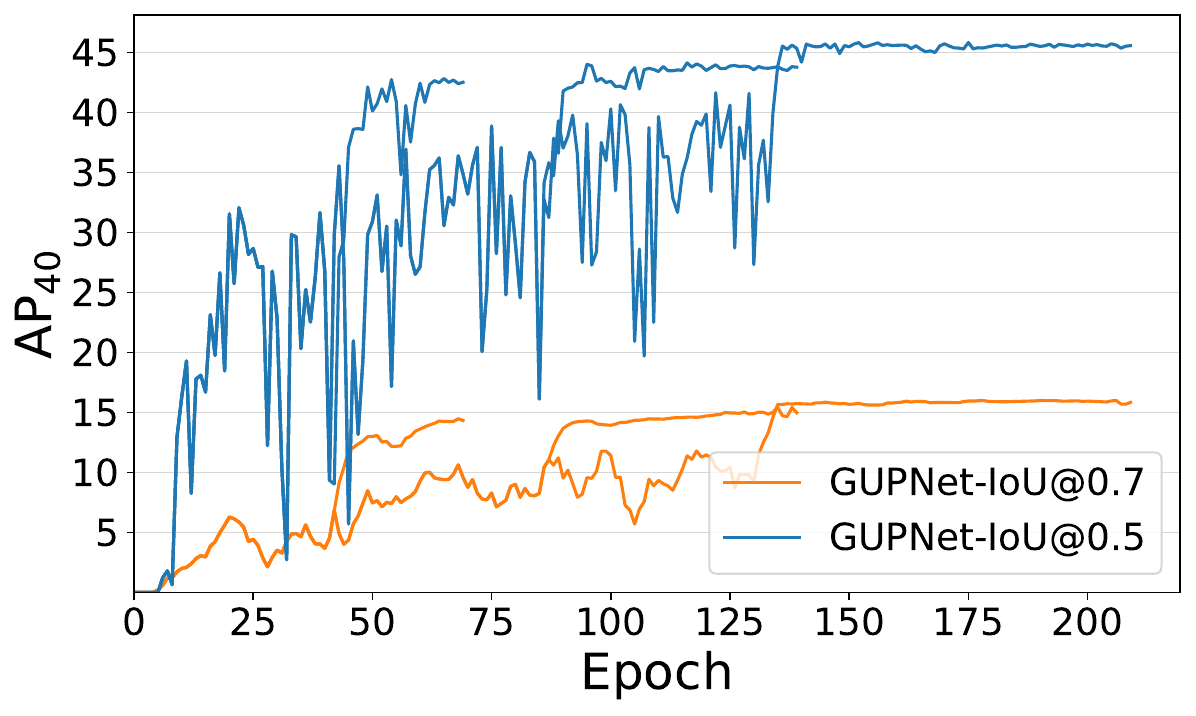}%
\vspace{-2ex}
\caption{
Learning curves on KITTI-3D without smoothing. 
}
\label{fig:schedule_curve_src}
\end{figure}

\subsection{Implementation of TIDE3D}
\label{appendix:implementation}

In the main paper, we introduce the design principles of TIDE3D, and there are some minor implementation differences on different datasets. 
Specifically, in KITTI-3D, TIDE3D conduct error diagnosis for three categories with three different settings separately. We directly compute the 3D IoU of detections and ground truths in the 3D space, instead of computing the BEV IoU and multiplying it by the 1D IoU at the height dimension. This design allows us to get more accurate 3D IoU for the objects whose roll angles and pitch angles are not zero (the roll angles and pitch angles of objects are always zero in KITTI-3D, but not necessarily for other datasets, such as nuScenes). In ScanNet V2, the error distribution is collected over all categories, \ie based on mAP. For nuScene, we compute both $\Delta \rm{mAP}$ and $\Delta \rm{NDS}$ for error diagnosis (we mainly focus on $\Delta \rm{mAP}$).
Besides, there is a hyper-parameter $t_{f}$ in TIDE3D, which is used to identify whether a false positive belongs to 'localization error' or 'background error'.  We set $t_{f}=0.1$ for 3D IoU based metrics and $t_{f}=5$ for center-distance based metrics.

\begin{figure*}[!t]
\centering
\subfloat[Pedestrian]{\includegraphics[width=3.3in]{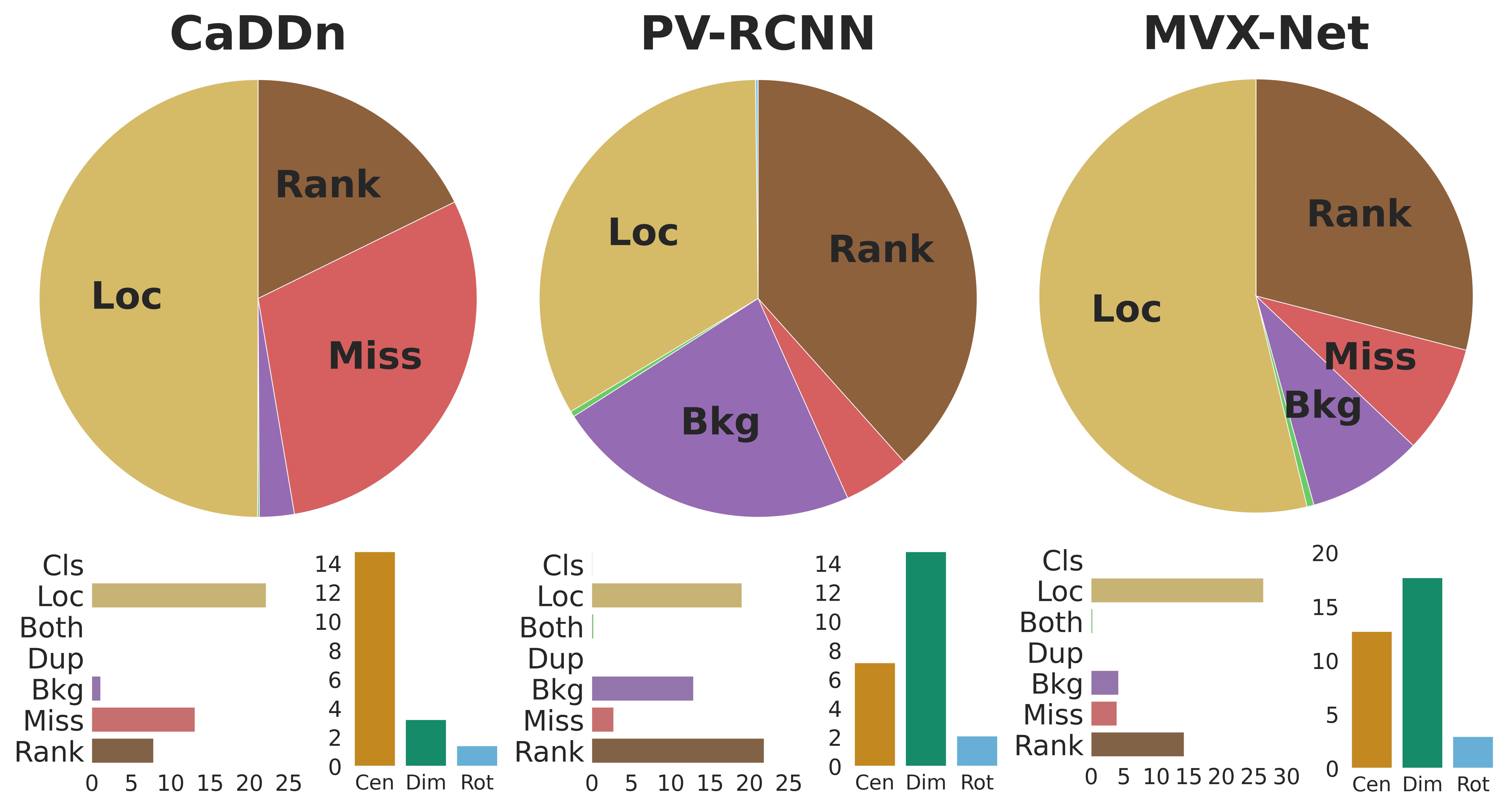}%
\label{fig:kitti_ped}}
\hfill{\color{black}\vrule}\hfill
\subfloat[Cyclist]{\includegraphics[width=3.3in]{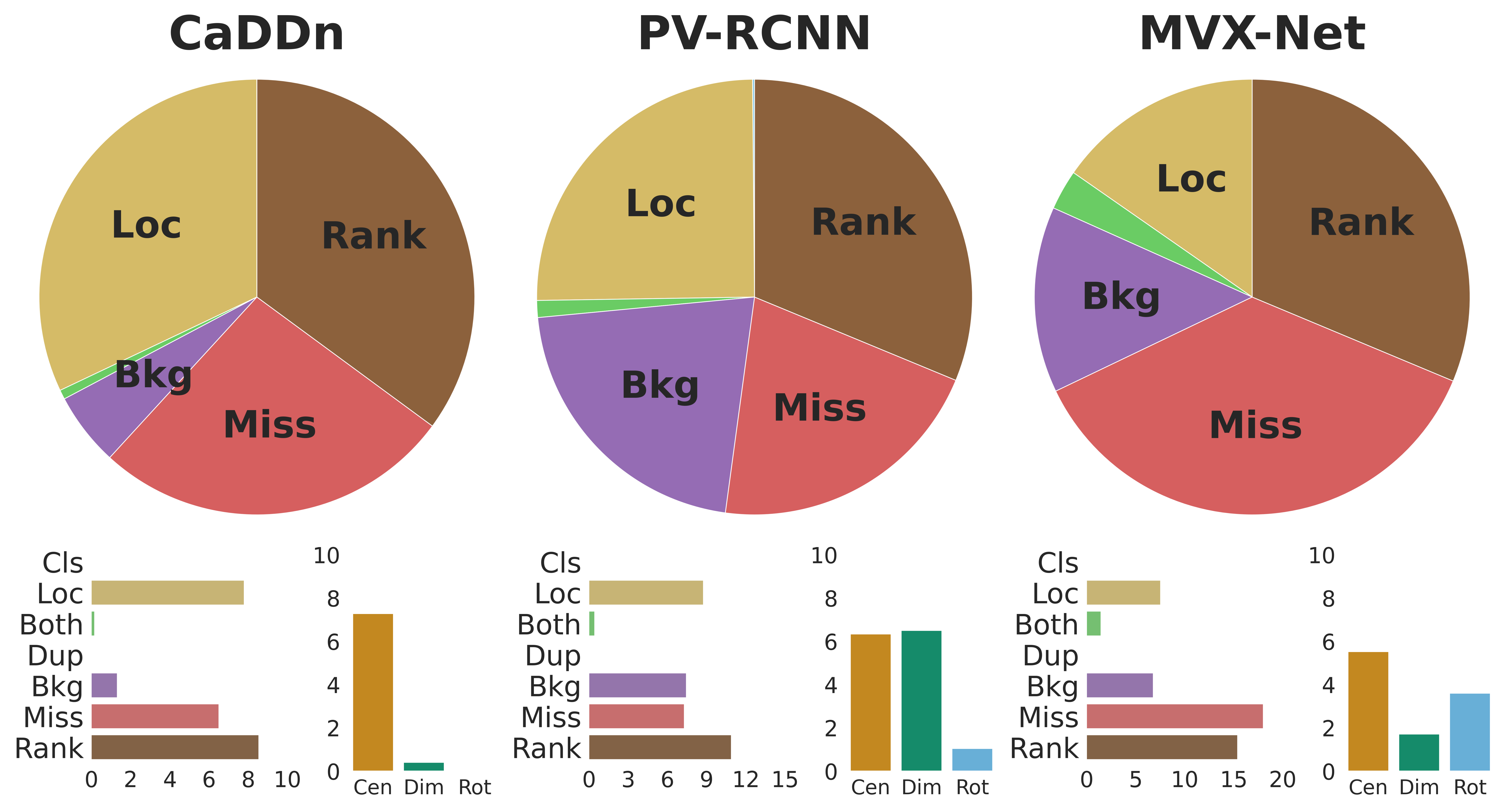}%
\label{fig:kitti_cyc}}
\caption{{\bf Example error diagnosis results.} 
We show the results of various models for the Pedestrian category and Cyclist category on KITTI-3D \emph{validation} set under the moderate setting with 0.5 IoU threshold.
Specifically, we select CaDDn~\cite{caddn} (image-based), PV-RCNN~\cite{pvrcnn} (LiDAR-based) and MVX-Net~\cite{mvxnet} (fusion-based) to explore the bottleneck of detectors with different modalities.
}
\label{fig:tide3d_kitti_ped_cyc}
\end{figure*}

\subsection{Training Curve}
\label{appendix:training_curve}

Figure \ref{fig:schedule_curve_src} shows the un-smoothed learning curve of GUPNet in KITTI-3D (Figure \ref{fig:schedule_curve}, right), and we can see that the performance fluctuates violently during training. Even in the later stage of training, the $\rm{AP}_{40}$ is still unstable (especially for the $1\times$ schedule). Meanwhile, we also observe the final performance fluctuates over multiple runs (similar to Table 20  in DEVIANT \cite{deviant}), and reporting mean values of multiple runs is recommended for future work.

\subsection{More TIDE3D Analysis}
\label{appendix:more_tide3d_analysis}

\begin{figure}[t]
\centering
\includegraphics[width=0.99\linewidth]{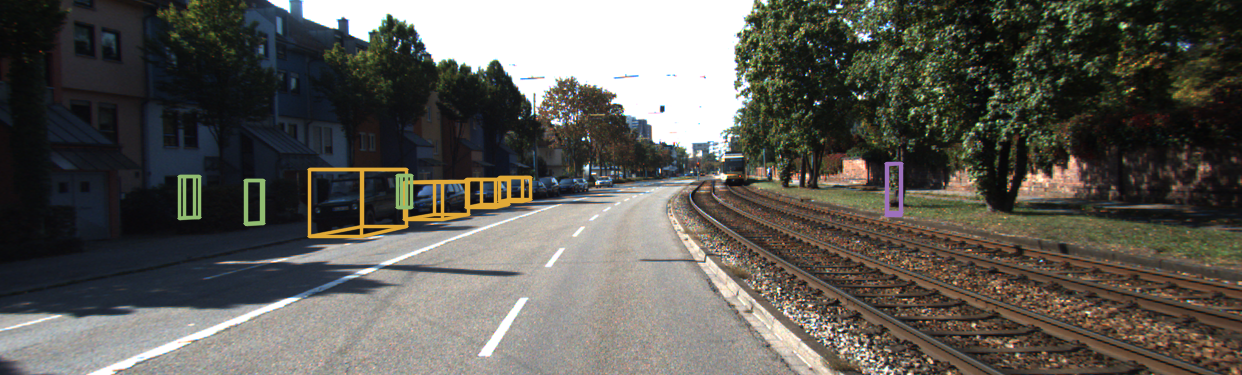}
\includegraphics[width=0.99\linewidth]{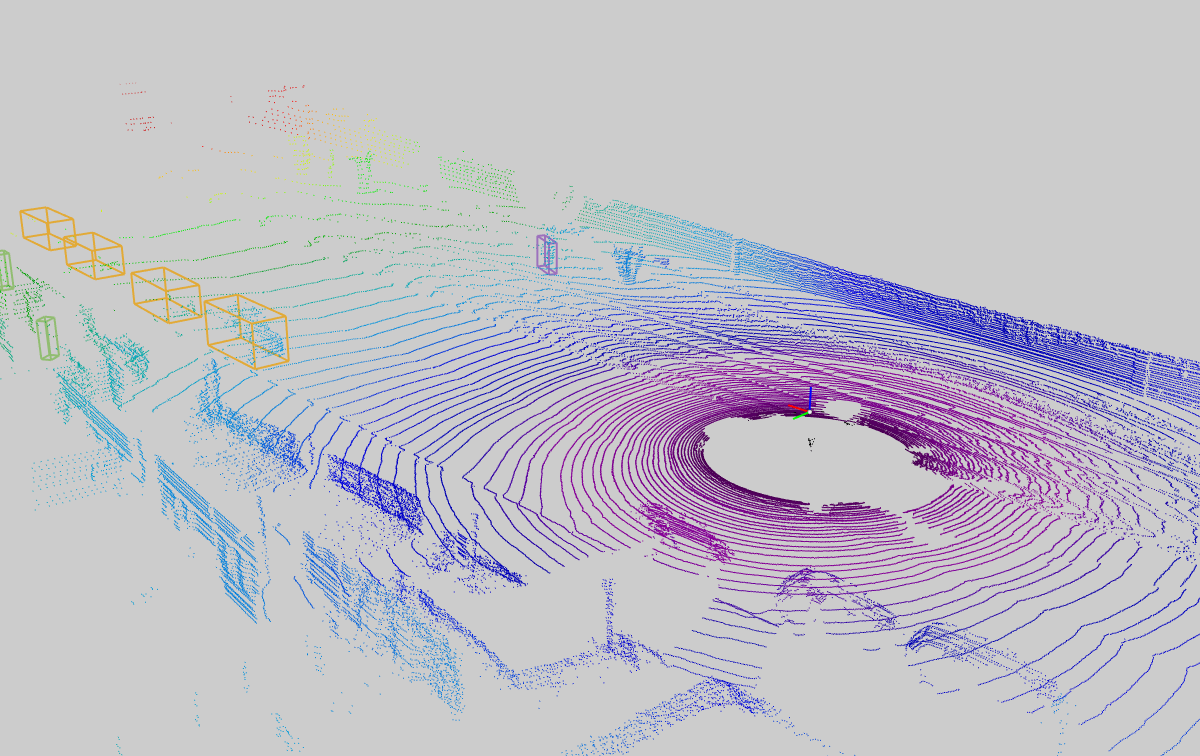}
\caption{An typical error case for LiDAR-based methods. A tree trunk is wrongly recognized as a cyclist due to the lack of color information. Cars, cyclists, and pedestrians are visualized with yellow, purple, and green boxes. Figures are copied from \cite{pseudolabeling}.}
\label{fig:errorcase}
\end{figure}

\noindent
{\bf Pedestrian and Cyclist.}
Here we show the error diagnosis result and discussion for the Pedestrian and Cyclist categories in Figure \ref{fig:tide3d_kitti_ped_cyc}. We can find that the major issue of the image-based model (CaDDn) is still the inaccurate localization. For the LiDAR-based model (PV-RCNN), the localization error is relatively small and mainly caused by the inaccurate estimation of dimension (instead of location). This is probably because LiDAR provides accurate location information but can only capture the surface of the objects. Besides, different from image-based methods, we can find that background error accounts for a large part of the whole error distribution. This is because LiDAR points lack color information, and the models base on them easily generate false positives when they meet objects with similar structures to the objects of interest. See Figure \ref{fig:errorcase} for a typical detection result.

\begin{figure}[!t]
\centering
\subfloat{\includegraphics[width=1.45in]{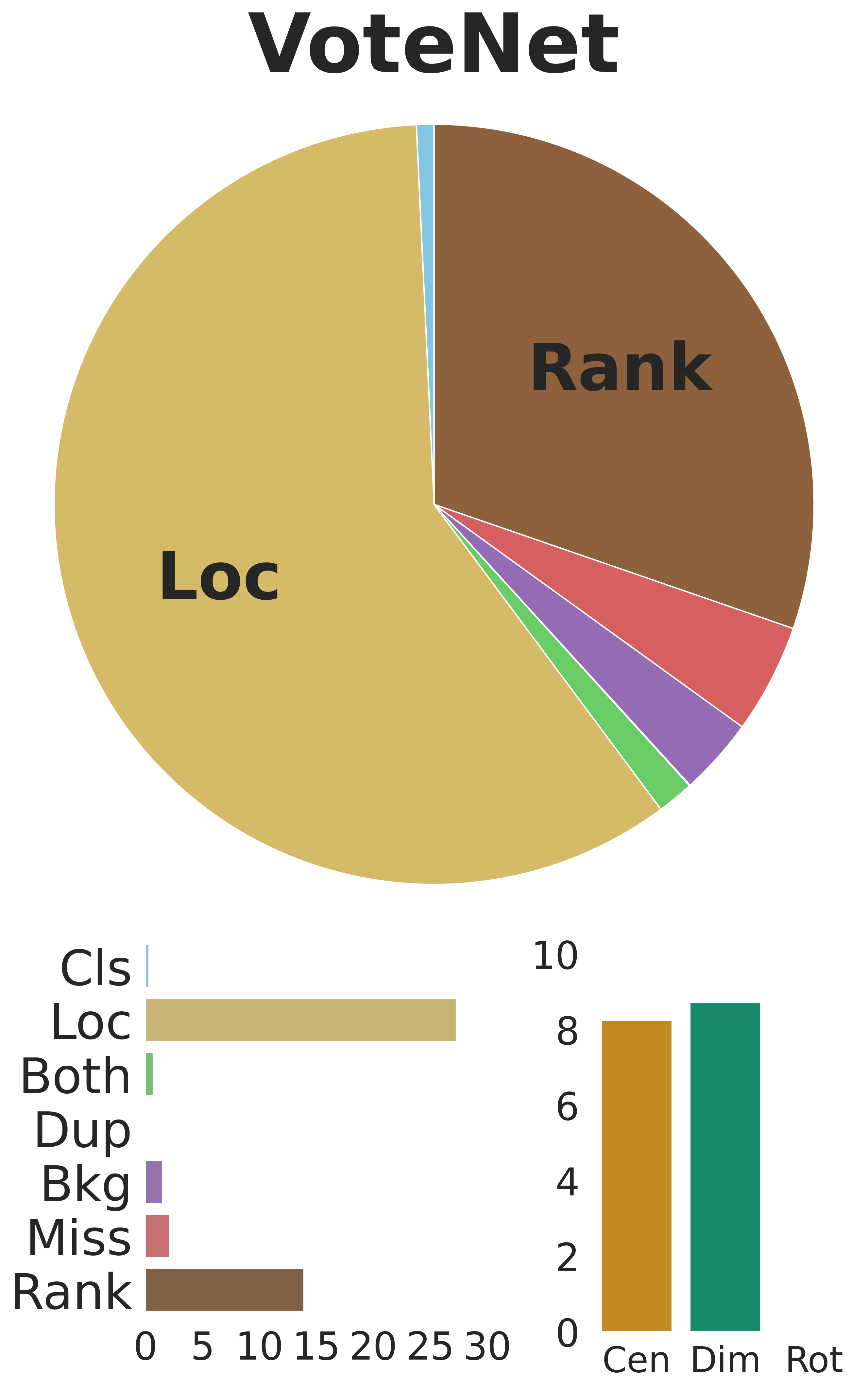}%
}
\hfill
\subfloat{\includegraphics[width=1.45in]{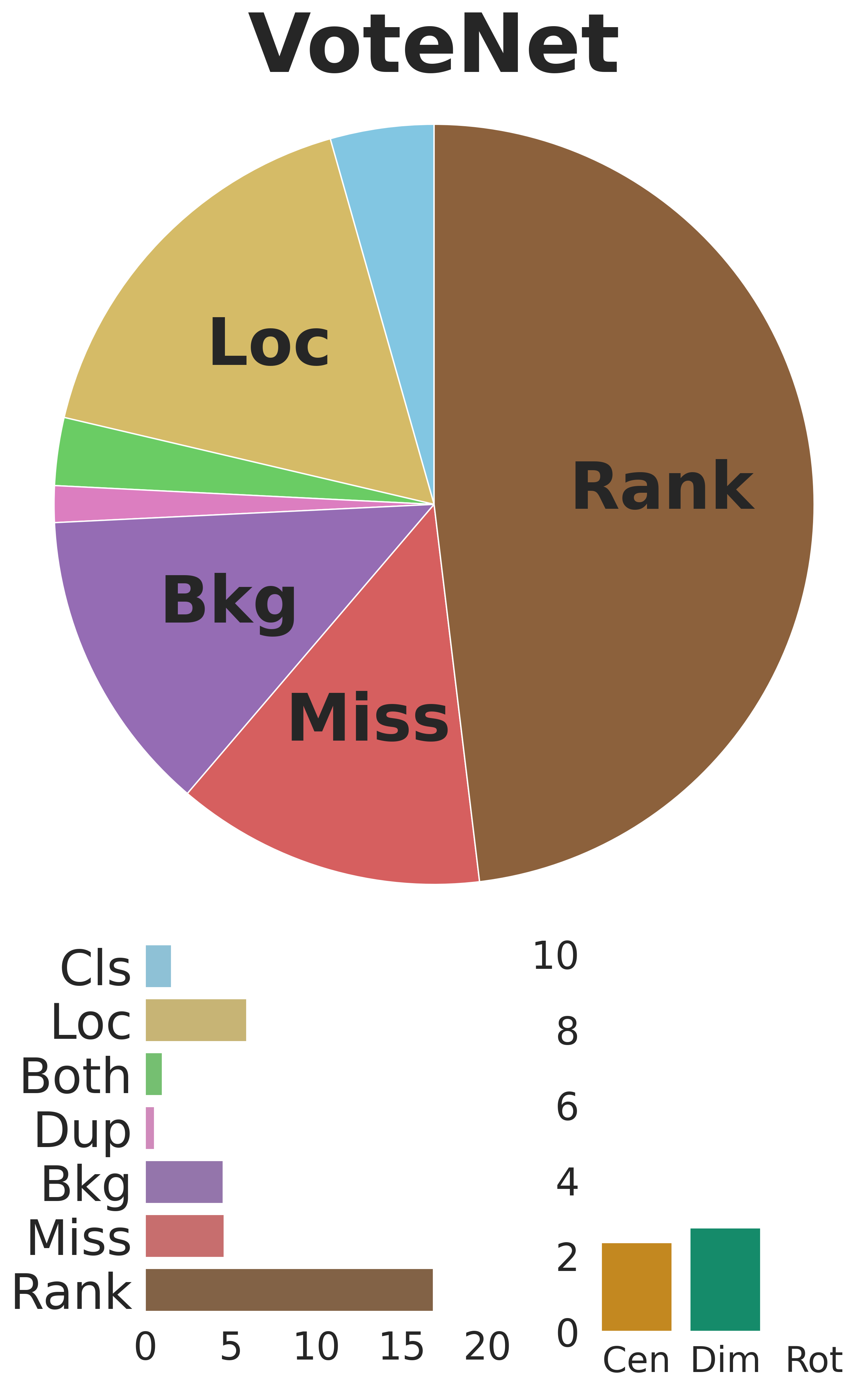}%
}

\caption{{\bf Example error diagnosis results of VoteNet on ScanNet V2 \emph{validation} set.} 
We show the results with 0.5 IoU threshold (\emph{left}) and 0.25 IoU threshold (\emph{right}) for all categories.}
\label{fig:tide3d_scannet}
\end{figure}

\noindent
{\bf Error diagnosis in ScanNet V2.} 
In addition to autonomous driving scenario, TIDE3D can also be used in indoor scenes. Here we use ScanNet v2 and VoteNet \cite{votenet}, which is a powerful detection model based on point cloud, as an example.
Figure~\ref{fig:tide3d_scannet} shows that VoteNet has different error distributions under different IoU thresholds.
Specifically, under a strict IoU threshold of 0.5, the model error is dominated by localization error which is caused by the inaccurate estimation of center and dimension (note the bounding box annotation in ScanNet V2 is axis-aligned and the model does not need to predict the rotation angle).
For a loose threshold of 0.25, the localization error is mitigated and the overall performance is limited by more factors, including ranking error, missing error, or background error. 

\begin{figure}[!t]
\centering
\subfloat{\includegraphics[width=1.45in]
{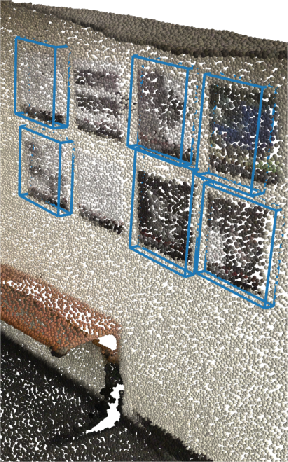}%
}
\hfill
\subfloat{\includegraphics[width=1.45in]{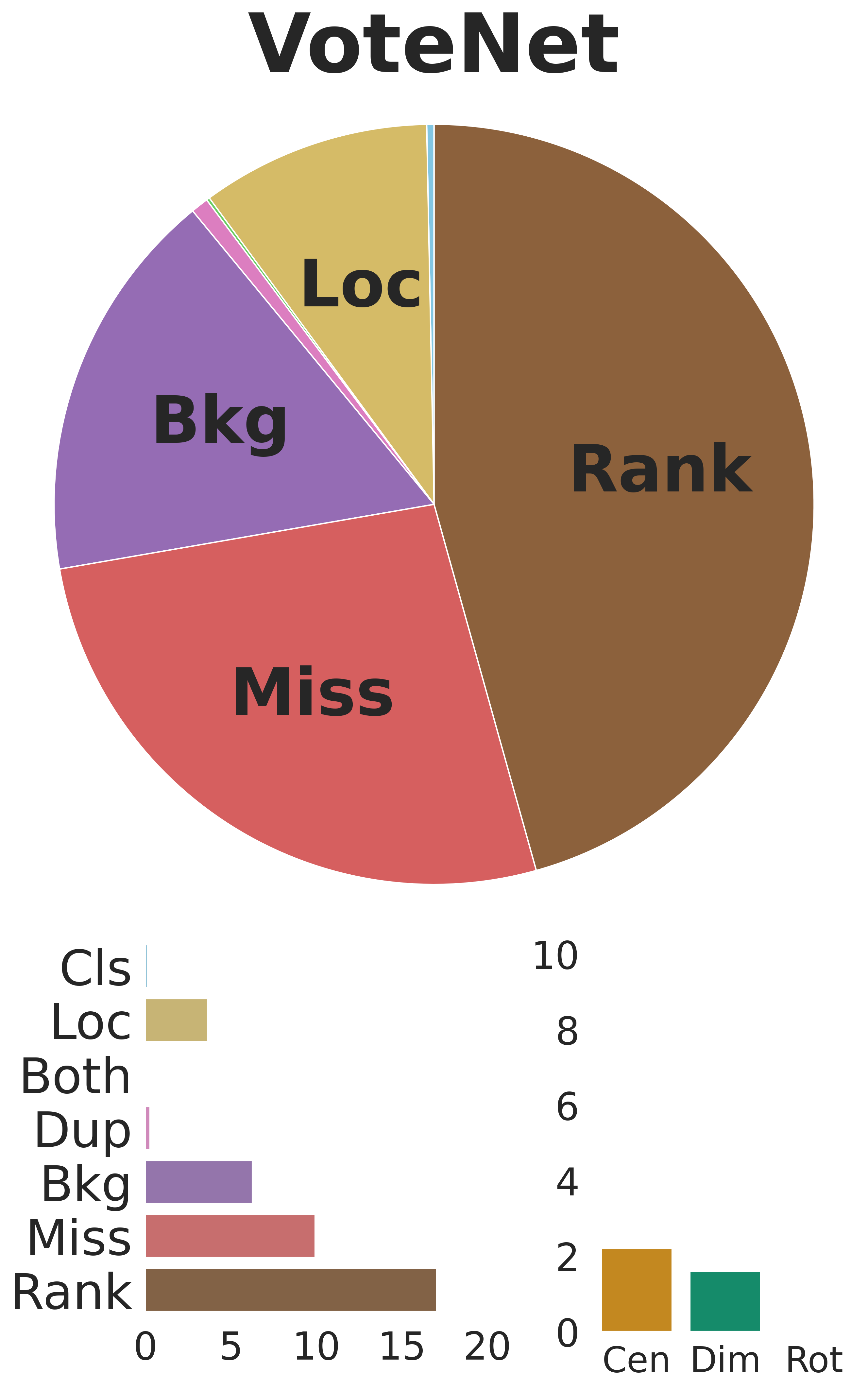}%
}
\caption{
{\bf Error diagnosis of VoteNet for `picture' category.} The error distribution is collected with the IoU threshold of 0.25 on ScanNet V2. A training sample with several `picture' items is provided to show the low quality of annotation, which is a major reason leading to the high background and missing errors. 
We use blue boxes to denote the ground truth items. Best viewed in color with zoom-in.}
\label{fig:scannet_picture}
\end{figure}

In order to analyze the diverse problems encountered in detecting different categories of objects, we can further analyze the AP of a specified single category. 
As an example, we look at the \textbf{`picture'} which is the worst performing category as shown in Figure~\ref{fig:scannet_picture}.
Since the pictures usually have small sizes and easily blend into the background point cloud of the wall, they are quite unrecognizable especially when there is no color information provided by the image data. 
Besides, incomplete annotations also confuse the detector's recognition of the pictures. Therefore, the main problems besides ranking error are missing error and background error for this category.

\subsection{Discussion about Training Recipes}
\label{appendix:augmentation}

We observe the common choices of training recipes show varying impacts on different models. Here we give some notes on model training:

\noindent
{\bf i.} The optimal training recipes are different for different models.
For example, the detection models with deformable convolutions \cite{dcn2}, such as \cite{smoke,monoflex}, require a lower learning rate (\eg $1.25e^{-4}$) for stable training, while other models with similar network architecture  \cite{monodle,gupnet} perform better at a larger one (\eg $1.25e^{-3}$). 

\noindent
{\bf ii.} Some data augmentations may have similar effects. For example, applying random crop (at a small-scale) and random shift separately can improve the accuracy, while applying both of them still get similar performance. This suggests that these two operations may work in a similar way, \ie applying geometric transformations to the images, thus making the models sensitive to the location of objects.

\noindent
{\bf iii.} For different datasets/metrics, the same augmentation may have different effects. For example, random crop works well in KITTI-3D dataset, while the performance change of applying this operation on nuScene is relatively inconspicuous. This is mainly caused by the difference in data size and evaluation metrics, and the custom training recipes for different application scenarios are required.

\noindent
{\bf iv.} Data augmentation is an effective way to improve the performance of detection models, even in large-scale datasets \cite{nuscenes,waymo}. However, it is difficult to align geometric changes in the 2D image plane and the 3D world space, resulting in limited augmentation strategies. The popular BEV pipeline allow us to conduct data augmentation in the BEV space, which is a key factor for the success of such methods. Meanwhile, this also indicates that the algorithm performance can be improved by designing data augmentation strategies in the future.

\subsection{Ranking Error in TIDE3D}
\label{appendix:more}
\begin{figure*}[!t]
\centering
\subfloat{\includegraphics[width=3.5in]{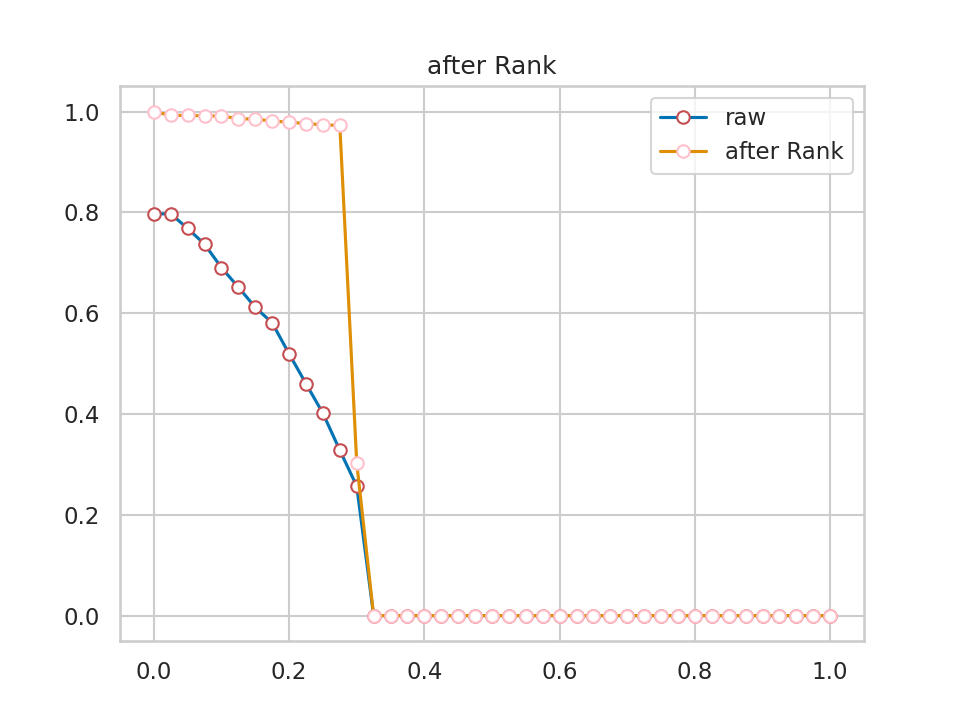}%
}
\subfloat{\includegraphics[width=3.5in]{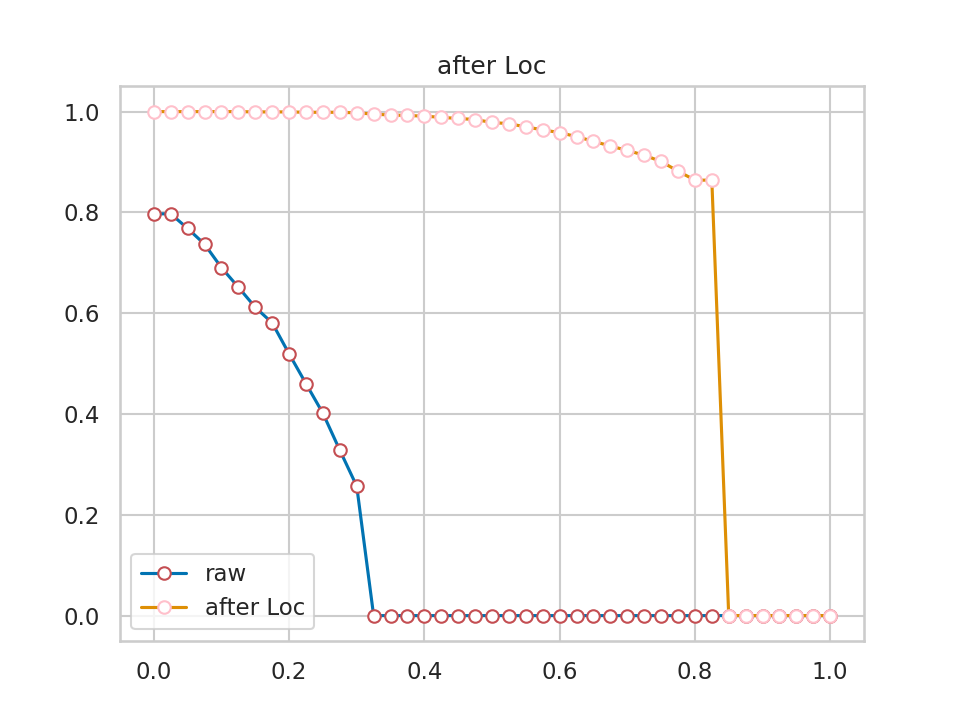}%
}

\caption{{\bf PR curves.} 
We show the PR curves before/after applying the ranking oracle ({\bf left}) and the localization oracle ({\bf right}). The baseline is GUPNet and the metric is $\rm{AP}_{40}$ on KITTI-3D \emph{validation} set under the moderate setting.}
\label{fig:prcurve}
\end{figure*}

We have shown that the ranking error, which is ignored in TIDE, is a major error type in 3D object detection systems. Different from other error types, ranking error involves multiple predictions, and fixing this type may affect other error types. For example, in Section \ref{sec:tide_analysis}, we show the effects of 3D confidence with TIDE3D and find that both the localization error and ranking error are significantly reduced (similar phenomenon is shown in Table 2 of TIDE \cite{tide}), which is caused by the complicated interactions between multiple predictions. We argue that, although improving the quality of localization and confidence may show similar numbers in localization error, they achieve this in a completely different way and can be further identified. Specifically, we present the Precision-Recall (PR) curves before/after applying localization and ranking oracles in Figure \ref{fig:prcurve}. We can find that the ranking (misalignment) oracle improves the AP by maximizing the precision at each recall level, while the localization oracle corrects the false positive to true positive. All the figures we presented will be given from TIDE3D.

\end{document}